\documentclass{article}

\PassOptionsToPackage{numbers, compress}{natbib}
\usepackage[preprint]{neurips_2026}

\usepackage[utf8]{inputenc}
\usepackage[T1]{fontenc}
\usepackage{hyperref}
\usepackage{url}
\usepackage{booktabs}
\usepackage{amsfonts}
\usepackage{amsmath}
\usepackage{amssymb}
\usepackage{nicefrac}
\usepackage{microtype}
\usepackage{xcolor}
\usepackage{graphicx}
\usepackage{multirow}

\newcommand{\best}[1]{\textcolor{red}{\textbf{#1}}}
\newcommand{\second}[1]{\textcolor{blue}{\underline{#1}}}

\title{GatedLinear: Adaptive Routing of Complementary Linear Bases for Time Series Forecasting}

\author{%
  Qitai Tan,\quad Ruiwen Gu,\quad Yilin Su,\quad Mo Li,\quad Xu Lin,\quad Xiao-Ping Zhang$^{\dagger}$ \\
  Shenzhen Key Laboratory of Ubiquitous Data Enabling, \\
  Shenzhen International Graduate School, Tsinghua University \\
  \texttt{tqt24@mails.tsinghua.edu.cn} \\
  \texttt{xpzhang@ieee.org}
}

\begin{document}

\maketitle
\renewcommand{\thefootnote}{\fnsymbol{footnote}}%
\footnotetext[2]{Corresponding author.}%
\setcounter{footnote}{0}

\begin{abstract}
Time series forecasting requires models to capture diverse, often mutually exclusive, temporal dynamics, from smooth trend continuation to nonstationary drift and strict phase-aligned recurrence. While recent deep learning models have improved accuracy, they typically force these diverse patterns through a single computational backbone governed by fixed algorithmic inductive biases (e.g., self-attention or spectral filtering). This single-mechanism approach often struggles with the profound heterogeneity of real-world series, where different variables and forecast horizons necessitate fundamentally different predictive treatments. To address this, we propose \textbf{GatedLinear}: a lightweight framework that frames forecasting as the \textbf{adaptive routing of complementary linear bases}. GatedLinear leverages a pool of \textbf{three specialized mechanisms}: a global trend-seasonal basis for smooth projection, a difference-based incremental basis for nonstationary drift, and a phase-aligned recurrence basis for explicit cyclic reuse. To dynamically orchestrate these distinct behaviors, we introduce a \textbf{Tri-Factorized Fusion Gate} that disentangles routing decisions into channel-specific preferences, horizon-aware offsets, and phase-indexed biases derived from known future time marks. This design allows the model to perform \textbf{highly granular, point-wise soft routing} across different predictive regimes without stacking computationally heavy neural modules. Experiments on standard benchmarks show that our method achieves state-of-the-art or highly competitive accuracy against recent complex foundational models, while offering explicitly interpretable routing patterns and operating with a substantially smaller parameter footprint.
\end{abstract}

\section{Introduction}

Time series forecasting is a fundamental problem in many real-world systems, including energy management, traffic control, weather prediction, finance, and industrial monitoring. The central challenge is not merely to fit historical observations, but to extrapolate future values under multiple temporal mechanisms. A future trajectory may continue a smooth trend, evolve through short-term increments, repeat a phase-aligned periodic pattern (i.e., exact historical recurring points such as 2:00 PM daily), or combine these behaviors in different proportions across variables and forecast horizons.

\begin{figure}[t]
  \centering
  \includegraphics[width=\linewidth]{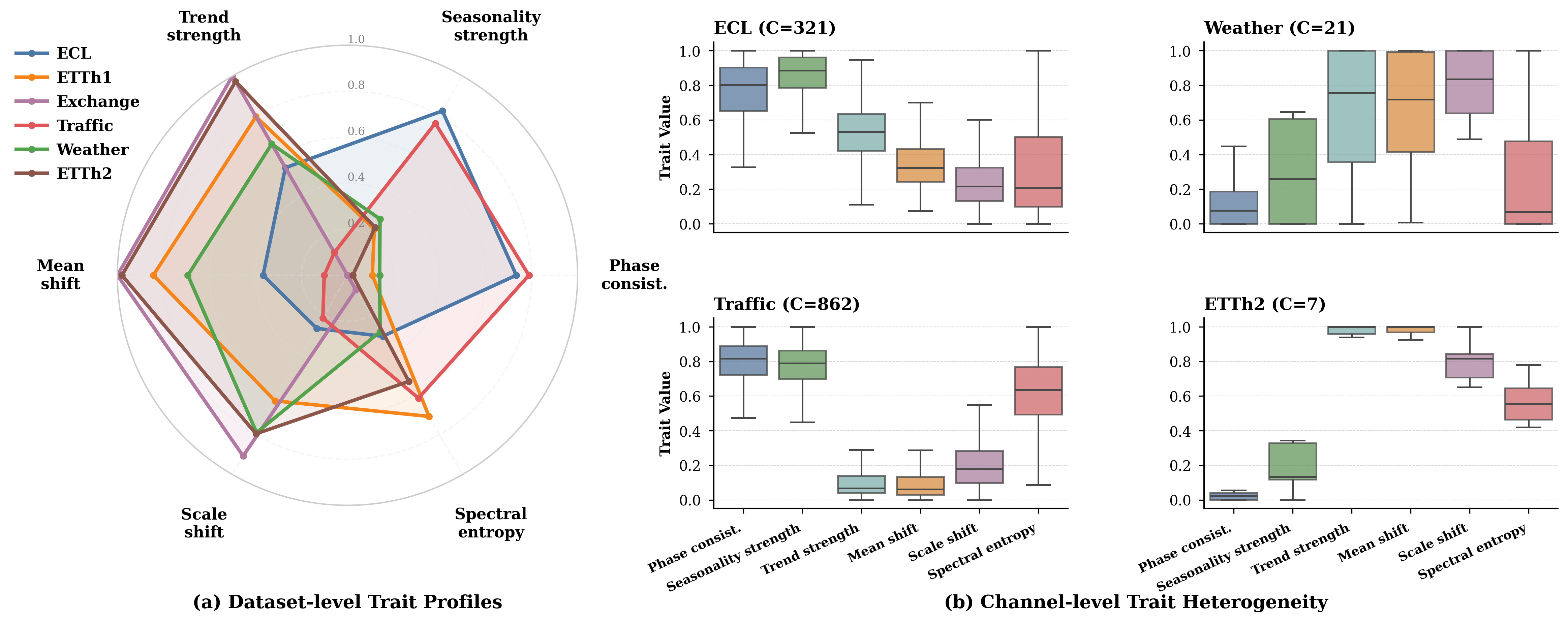}
  \caption{Trait heterogeneity in time series forecasting benchmarks. (a) Different datasets exhibit distinct profiles across temporal traits, suggesting that no single forecasting mechanism is uniformly appropriate across datasets. (b) Even within one dataset, variable channels can have broad trait distributions, indicating substantial channel-level heterogeneity. The six temporal traits are defined in the Appendix~\ref{app:trait-gate-correlation}.}
  \label{fig:intro-traits}
\end{figure}

Previously, time series forecasting predominantly relied on classical statistical models, such as ARIMA \citep{ARIMA}, vector autoregression (VAR) \citep{VAR}, and exponential smoothing (ETS) \citep{ETS}. With the rise of deep learning, early neural approaches further advanced the field, including RNN-based DeepAR \citep{salinas2020deepar}, LSTNet \citep{lai2018modeling}, and SegRNN \citep{lin2025segrnn}, as well as CNN-based TCN \citep{bai2018empirical} and SCINet \citep{liu2022scinet}. More recently, the continuing advancement of deep learning has derived several prominent branches that steer the development of the field. Linear and MLP-based methods (e.g., DLinear \citep{zeng2023transformers}, TimeMixer \citep{wangtimemixer}, TimeAlign \citep{hu2025bridging}) pursue lightweight forecasting by emphasizing trend--seasonal decomposition, multi-scale mixing, or distribution-aware alignment. Transformer-based techniques (e.g., Autoformer \citep{wu2021autoformer}, PatchTST \citep{patchtst}, iTransformer \citep{liuitransformer}, PhaseFormer \citep{niu2025phaseformer}) excel at modeling global dependencies through self-attention mechanisms, albeit at the cost of high computational complexity. Beyond these mainstreams, a line of methods tailored for specific temporal dynamics has also received significant attention, where frequency- and periodicity-oriented models (e.g., FilterNet \citep{yi2024filternet}, TimesNet \citep{wutimesnet}, FreqCycle \citep{zhang2026freqcycle}, TQNet \citep{tqnet}) explicitly exploit spectral structures and periodic recurrence to capture repeating components.

\begin{figure}[t]
  \centering
  \includegraphics[width=\linewidth]{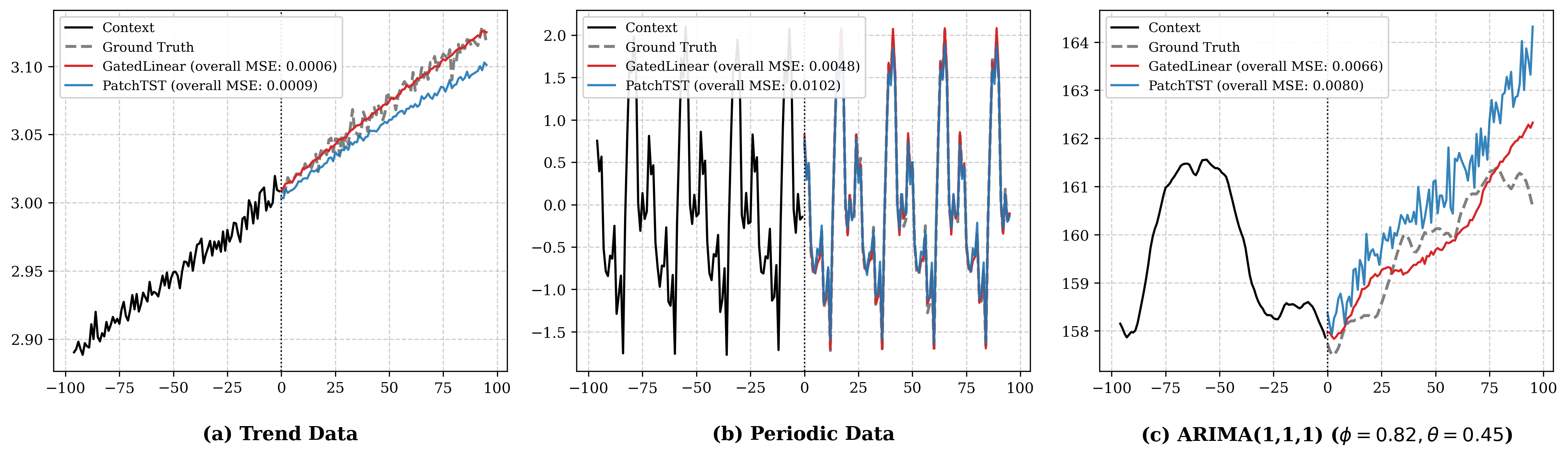}
  \caption{Performance comparison between PatchTST and GatedLinear on data with distinct temporal characteristics: (a) Trend Data with smooth continuous evolution, (b) Periodic Data with strict phase-aligned recurrence, and (c) ARIMA(1,1,1) data with nonstationary drift.}
  \label{fig:intro-synthetic}
\end{figure}

Despite this progress, the representation power of many advanced methods is heavily constrained by their underlying algorithmic inductive biases (e.g., self-attention inherently prefers smooth interpolation, while frequency filters assume stable periodicities). Rather than adaptively altering their forecasting mechanism, these architectures force all inputs through a unified computational backbone. However, real-world time series exhibit profound heterogeneity that often involves mutually exclusive temporal dynamics. As shown in Figure~\ref{fig:intro-traits}, datasets differ substantially in trend strength, exact periodicity, and local drift, and even variables within the same dataset can follow distinct temporal regimes. SynTSBench \citep{tansyntsbench} further supports this view by constructing synthetic datasets with programmable temporal patterns and showing that few deep forecasting models perform uniformly well across all pattern types. A stable daily pattern explicitly benefits from recurring phase-aligned reuse---directly looking up the corresponding phase from historical cycles; a nonstationary drifting signal inherently requires incremental evolution; and a smooth trajectory is better handled by direct trend-seasonal projection. Relying on a single mechanism often leads to compromised predictions. For instance, Figure~\ref{fig:intro-synthetic} illustrates that a representative Transformer, PatchTST, underfits strict trends, over-smooths exact cyclic spikes, and lags behind local nonstationary drift. These observations motivate a more dynamic forecasting paradigm: instead of expanding a single neural backbone, we propose an adaptive framework that routes each forecast point, identified by its variable, horizon, and future phase index, to an appropriate mixture of tailored, complementary linear mechanisms.

Motivated by this view, we propose \textbf{GatedLinear}, a lightweight framework that formulates time series prediction as adaptive routing over a pool of three complementary linear bases. The first basis performs decomposed direct mapping over seasonal and trend components, capturing smooth continuation patterns. The second basis predicts future increments and integrates them forward from the last observation, targeting nonstationary but locally predictable dynamics. The third basis constructs a phase-aligned recurrence template from recent historical cycles and refines it with a linear residual, explicitly exploiting periodic reuse.

To combine these bases, we introduce a \textbf{Tri-Factorized Fusion Gate}. Instead of assigning a fixed branch weight to an entire dataset or sample, the gate produces weights for each forecast point and variable. Its logits are factorized into a \textbf{channel-specific branch preference}, a \textbf{horizon-channel correction}, and a \textbf{future-phase-indexed correction} whose index is derived from decoder-side time marks. This design keeps the forecasting branches linear and parameter-efficient, while enabling fine-grained, interpretable selection across horizons, variables, and periodic phases.

\begin{figure}[t]
  \centering
  \includegraphics[width=\linewidth]{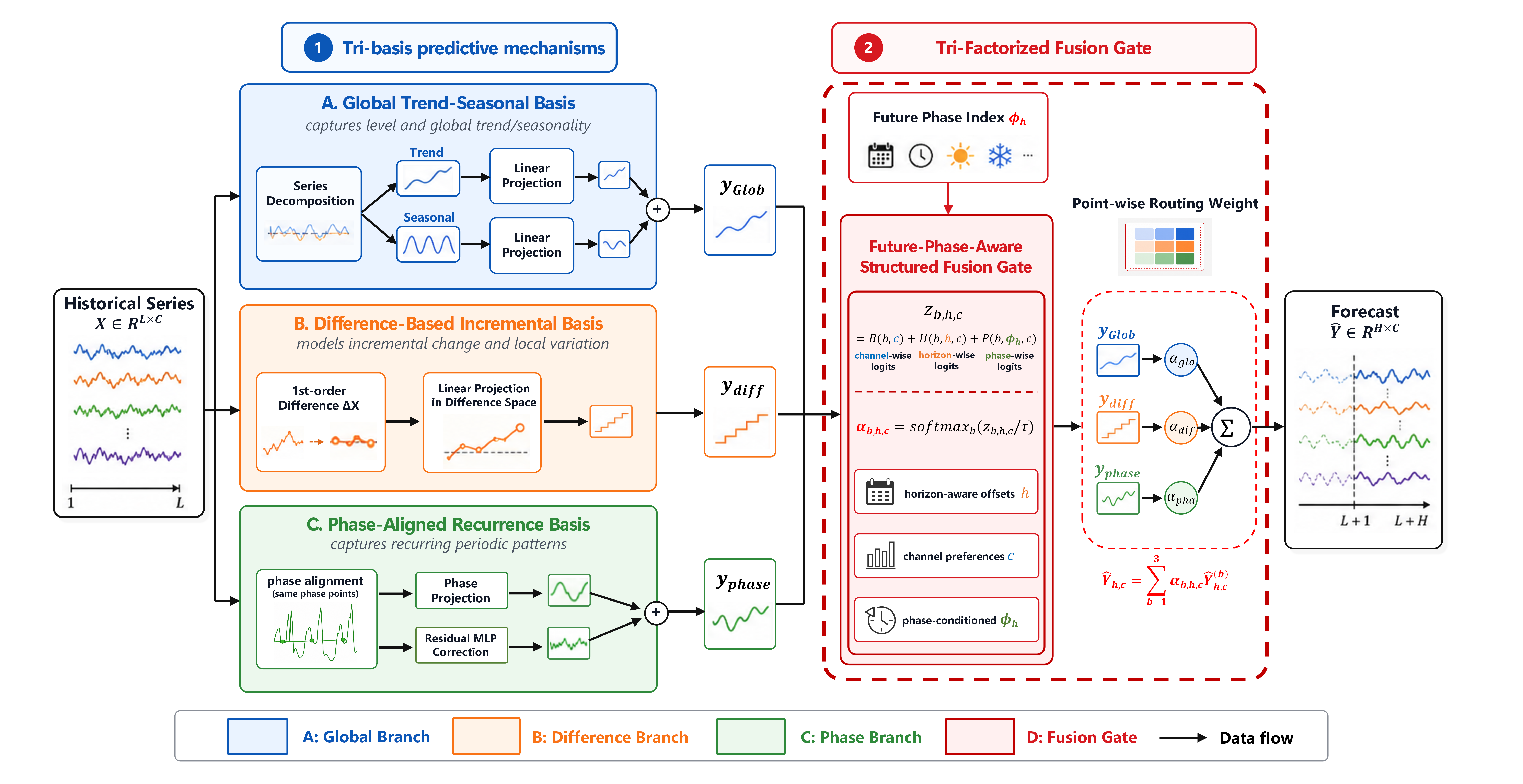}
  \caption{Overview of the proposed GatedLinear framework. The input sequence is processed by three complementary linear forecasting bases. A Tri-Factorized Fusion Gate uses horizon, channel, and future phase indices to select among the three bases for each future point.}
  \label{fig:framework}
\end{figure}

Figure~\ref{fig:framework} illustrates the overall architecture of the proposed framework, including the three forecasting bases and the Tri-Factorized Fusion Gate.

Our contributions are summarized as follows:
\begin{itemize}
  \item We propose \textbf{GatedLinear}, a tri-basis linear forecasting framework that explicitly models a pool of three complementary predictive mechanisms: decomposed direct projection, difference-based incremental evolution, and phase-aligned periodic recurrence. The framework naturally addresses dataset-level and channel-level temporal heterogeneity via dynamic routing without stacking heavy neural modules.
  \item We introduce a \textbf{Tri-Factorized Fusion Gate} that enables forecast-point-level, channel-wise, and phase-indexed soft blending. By factorizing gate logits into channel base tendencies, horizon offsets, and future-phase-indexed biases, the model precisely manages the basis mixture while preserving interpretability.
  \item We demonstrate that adaptive routing of basic linear mechanisms allows a lightweight model to achieve state-of-the-art or highly competitive performance on standard forecasting benchmarks, suggesting that effective dynamic blending is as critical as increasing architectural complexity.
\end{itemize}

\section{Problem formulation}

Given a multivariate time series $\mathbf{X}_{t-L+1:t}\in\mathbb{R}^{L\times C}$ with $L$ historical observations and $C$ variables, the goal is to predict future values $\hat{\mathbf{Y}}_{t+1:t+H}\in\mathbb{R}^{H\times C}$, where $H$ denotes the prediction length. The forecasting task can be formalized as learning a mapping
\begin{equation}
  f:\mathbb{R}^{L\times C}\rightarrow\mathbb{R}^{H\times C},
  \qquad
  \hat{\mathbf{Y}}_{t+1:t+H}=f(\mathbf{X}_{t-L+1:t}).
\end{equation}
Our goal is to design a lightweight forecasting function $f$ that can adapt its predictive mechanism across variables, forecast steps, and future temporal phases. For notational simplicity, we drop the global timestamp $t$ in the following sections and locally index the historical window as $\mathbf{X}_{1:L}$ and the future horizon as $\hat{\mathbf{Y}}_{1:H}$.

\section{Methodology}

This section describes the three forecasting bases and the structured fusion gate in detail.

\subsection{Tri-basis linear forecasting}

Our model instantiates the forecasting function $f$ using three complementary linear bases. Each basis generates a candidate future sequence $\hat{\mathbf{Y}}^{(k)} \in \mathbb{R}^{H \times C}$ for $k \in \{1, 2, 3\}$, corresponding to the Global Trend-Seasonal Basis, Difference-Based Incremental Basis, and Phase-Aligned Recurrence Basis, respectively. A fusion gate simultaneously yields mixing weights $\mathbf{W} \in \mathbb{R}^{H \times C \times 3}$, such that $\sum_{k=1}^{3} \mathbf{W}_{h,c,k} = 1$ for all future steps $h \in \{1, \dots, H\}$ and channels $c \in \{1, \dots, C\}$. The final sequence prediction is a convex combination of the branches:
\begin{equation}
  \hat{\mathbf{Y}}_{h,c} = \sum_{k=1}^{3} \mathbf{W}_{h,c,k} \hat{\mathbf{Y}}^{(k)}_{h,c}.
\end{equation}
Here, $\hat{\mathbf{Y}}_{h,c}$ denotes the forecasted value at targeted future step $h$ for variable $c$, and $\mathbf{W}_{h,c,k}$ is the weight confidently assigned to basis $k$ for that specific point.

\paragraph{Global Trend-Seasonal Basis ($k=1$).}
The global basis is designed to learn smooth trend and seasonal structures that can be directly extrapolated over the forecast horizon. Following decomposition-based forecasting paradigms, the historical input is split into seasonal and trend components via moving-average operations:
\begin{equation}
  \mathbf{X}^{s}, \mathbf{X}^{t} = \mathrm{Decomp}(\mathbf{X}),
\end{equation}
where $\mathbf{X}^{s}, \mathbf{X}^{t} \in \mathbb{R}^{L \times C}$ isolate local cyclical fluctuations and global trends from the input $\mathbf{X}$. Two linear mappings then project these components from the historical length $L$ directly to the entire future horizon $H$:
\begin{equation}
  \hat{\mathbf{Y}}^{(1)}_{:, c} = \mathcal{L}_{s}(\mathbf{X}^{s}_{:, c}) + \mathcal{L}_{t}(\mathbf{X}^{t}_{:, c}), \quad \forall c \in \{1,\dots,C\},
\end{equation}
where $\mathcal{L}_{s}, \mathcal{L}_{t}: \mathbb{R}^{L} \to \mathbb{R}^{H}$ are learnable temporal affine layers operating over individual channels.

\paragraph{Difference-Based Incremental Basis ($k=2$).}
The difference basis targets nonstationary series with level shifts or persistent local drift, where predicting changes can be more stable than predicting absolute levels. We compute the first-order differences of the input series:
\begin{equation}
  \Delta\mathbf{X}_{\tau, c} = \mathbf{X}_{\tau, c} - \mathbf{X}_{\tau-1, c}, \quad \forall \tau \in \{2, \dots, L\},
\end{equation}
with $\Delta\mathbf{X}_{1, c} = 0$. Using a temporal linear projection $\mathcal{L}_{d}: \mathbb{R}^{L} \to \mathbb{R}^{H}$, this branch forecasts all future increments sequentially. The target future levels $\hat{\mathbf{Y}}^{(2)}$ are explicitly reconstructed via cumulative summation, integrated starting strictly from the last historically observed value $\mathbf{X}_{L, c}$:
\begin{equation}
  \hat{\mathbf{Y}}^{(2)}_{h,c} = \mathbf{X}_{L,c} + \sum_{i=1}^{h} \left[ \mathcal{L}_{d}(\Delta\mathbf{X}_{:, c}) \right]_{i},
\end{equation}
where $i$ indexes the forecasted increments up to step $h$, so future levels are reconstructed by accumulating predicted changes from the last observed value. By explicitly integrating predicted increments from the exact last historical observation $\mathbf{X}_{L,c}$, this basis inherently avoids the unconditional bias caused by severe level shifts. A detailed theoretical analysis of its optimality in non-stationary drifting environments is provided in Appendix~\ref{app:theoretical-analysis}.

\paragraph{Phase-Aligned Recurrence Basis ($k=3$).}
The phase-aligned basis complements the other two bases by mapping fine-grained periodic details from recent cycles to their corresponding future phases. For a given time series with characteristic base period $P$, assuming the historical window is sufficiently long ($L \ge K \cdot P$), we summarize the last $K$ complete periods into a unified historical template $\mathbf{T} \in \mathbb{R}^{P \times C}$. For each recurring phase $r \in \{0, \dots, P-1\}$:
\begin{equation}
  \mathbf{T}_{r, c} = \frac{1}{K} \sum_{k=1}^{K} \mathbf{X}_{L - K P + (k-1)P + r, c}.
\end{equation}
This exact template is subsequently tiled forward across the desired prediction length $H$ to yield the unparameterized baseline sequence $\mathbf{B}^{(p)} \in \mathbb{R}^{H \times C}$. We fine-tune this cyclical projection by supplementing an additive linearly modeled deformation residual $\mathcal{L}_{p}(\mathbf{X})$:
\begin{equation}
  \hat{\mathbf{Y}}^{(3)}_{:, c} = \mathbf{B}^{(p)}_{:, c} + \mathcal{L}_{p}(\mathbf{X}_{:, c}).
\end{equation}
The template provides a same-phase forecast prior, while the residual absorbs mild deviations from exact recurrence.

Notably, all three bases share the exact same structural simplicity: they are constructed solely from fundamental mathematical operations (difference, moving-average, phase alignment) combined with one-layer temporal linear mappings, entirely free of heavy neural modules.

\subsection{Tri-Factorized Fusion Gate}

To invoke the bases adaptively, we equip our framework with a structured gating mechanism yielding the mixing coefficients $\mathbf{W}$. We designate this as a "\textit{Tri-Factorized Fusion Gate}" because the gate separates three lookup contexts: the variable channel, the forecast horizon, and the future phase index associated with the target timestamp. For example, under an hourly daily period, the gate can treat a noon forecast differently from a midnight forecast even when the two targets have the same horizon length. This fine-grained, localized alignment enables point-wise mechanism selection.

Accordingly, for future step $h \in \{1,\dots,H\}$, given channel $c$, and candidate basis $k$, the unnormalized preference logit $z_{h,c,k}$ is structured into three disjoint learnable parameters:
\begin{equation}
  z_{h,c,k} = a_{c,k} + u_{h,c,k} + v_{\phi_h, c, k},
\end{equation}
with respective meanings assigned as follows:
\begin{itemize}
  \item $a_{c,k} \in \mathbb{R}$ conveys a persistent channel-specific base tendency, enabling varying dominances among different monitored variables.
  \item $u_{h,c,k} \in \mathbb{R}$ implements a horizon-dependent scalar offset, reflecting that smooth direct bases inherently excel farther out while increment predictions thrive primarily under tight proximities.
  \item $v_{\phi_h,c,k} \in \mathbb{R}$ is a phase-indexed selection bias, selected by the future phase index $\phi_h \in \{0, \dots, P-1\}$ associated with the target timestamp. This index is obtained from known decoder-side time marks when available, or from the horizon offset modulo the period otherwise.
\end{itemize}

These structured logits are ultimately normalized into bounded fusion probabilities via a parametric softmax distribution ruled by a temperature hyperparameter $\tau$:
\begin{equation}
  \mathbf{W}_{h,c,k} = \frac{\exp(z_{h,c,k}/\tau)}{\sum_{j=1}^{3} \exp(z_{h,c,j}/\tau)}.
\end{equation}
Reducing $\tau$ intrinsically prevents ambiguous blending and compels the gate to isolate the most reliable basis dominantly.

Remarkably, our constructed framework entirely circumvents weighty recursive networks or excessive cross-attention modules. It establishes a compact selection scheme governed by structured lookups over channel, horizon, and phase index.

\subsection{Training objective and complexity}

For forecasting, we optimize the network using the standard Mean Squared Error (MSE) loss over a training batch:
\begin{equation}
  \mathcal{L}_{\mathrm{forecast}}
  = \frac{1}{B \times H \times C} \sum_{b=1}^{B} \sum_{h=1}^{H} \sum_{c=1}^{C} \left( \hat{\mathbf{Y}}_{b,h,c} - \mathbf{Y}_{b,h,c} \right)^2,
\end{equation}
where $B$ designates the batch size, and $\mathbf{Y}_{b,h,c}$ represents the ground-truth future values for a specific sequence in the batch.

\paragraph{Complexity analysis.}
Our tri-basis framework maintains a strictly lightweight profile, deliberately circumventing the quadratic overheads predominantly associated with self-attention architectures. We itemize complexity in two integral dimensions:
\begin{itemize}
  \item \textbf{Parameter complexity:} Under the common channel independence setting where weights are shared across variates, each linear mapping branch ($\mathcal{L}_s, \mathcal{L}_t, \mathcal{L}_d, \mathcal{L}_p$) utilizes exactly $\mathcal{O}(L \times H)$ weights, accumulating to $\mathcal{O}(L \times H)$ parameters in total. The structured gate introduces disjoint lightweight parameter matrices encompassing exactly $3 \times C$ for basis bias, $3 \times H \times C$ for horizon offset, and $3 \times P \times C$ for specific phase conditioning. Consequently, the overall model parameter count scales as $\mathcal{O}(LH + C(H+P))$, keeping the size notably bounded and effectively bypassing intensive MLP topologies.
  \item \textbf{Computational complexity:} During the forward pass, extracting components and applying temporal matrix projections inherently requires $\mathcal{O}(L \times H)$ operations per individual channel. Iterating throughout the entire batch results in a theoretical time complexity scaling compactly strictly as $\mathcal{O}(B \times C \times L \times H)$. The final gating fusion seamlessly computes simple additions and table lookups requiring negligible linear runtime strictly scaling as $\mathcal{O}(B \times C \times H)$. Contrastingly, conventional global Transformers enforce prohibitive $\mathcal{O}(B \times C \times L^2)$ or unified sequence computational boundaries. Our decoupled design ensures $\mathcal{O}(BCLH)$ deterministic, lightning-fast end-to-end execution.
\end{itemize}

\section{Experiments}

\subsection{Experimental setup}

\paragraph{Datasets.}
We evaluate on standard forecasting benchmarks, including ETTm1, ETTm2, ETTh1, ETTh2, Electricity, Traffic, Weather, and Exchange \citep{zhou2021informer,wu2021autoformer,lai2018modeling}. Unless otherwise stated, each dataset is split into train/validation/test subsets with a ratio of 7:1:2. We use an input length of $L=336$ and evaluate prediction lengths $H\in\{48,96,192,336\}$. For datasets with clear daily or weekly periodicity, we set the period length according to the data frequency; otherwise, we use a validation-selected period.

\paragraph{Baselines.}
We compare against the same representative models reported in Table~\ref{tab:main-results}: DLinear \citep{zeng2023transformers}, TimeMixer \citep{wangtimemixer}, MixLinear \citep{ma2024mixlinear}, PaiFilter \citep{yi2024filternet}, PatchTST \citep{patchtst}, PhaseFormer \citep{niu2025phaseformer}, TQNet \citep{tqnet}, TimeAlign \citep{hu2025bridging}, and iTransformer \citep{liuitransformer}. These baselines cover decomposition-based linear models, multi-scale MLP forecasting, patch- and phase-based temporal modeling, frequency filtering, temporal-query mechanisms, distribution-aware alignment, and inverted Transformer architectures. Detailed descriptions and source repositories are provided in the appendix.

\paragraph{Implementation details.}
All experiments are implemented in PyTorch 2.6.0 based on the Time-Series-Library training pipeline and conducted on a single NVIDIA A800 GPU. We train GatedLinear with Adam using mean squared error (MSE) loss, and report mean absolute error (MAE) and MSE as evaluation metrics. Further implementation details are provided in Appendix~\ref{app:implementation-details}.

\subsection{Main results}

\begin{table*}[t]
  \caption{Main forecasting results (averaged over prediction lengths $H\in\{48,96,192,336\}$ with input length $L=336$). The best results are shown in \best{red bold}, and the second-best in \second{blue underline}.}
  \label{tab:main-results}
  \centering
  \setlength{\tabcolsep}{3pt}
  \resizebox{\textwidth}{!}{%
  \begin{tabular}{c|cc|cc|cc|cc|cc|cc|cc|cc|cc|cc}
    \toprule
    \multirow{2}{*}{\textbf{Dataset}} & \multicolumn{2}{c|}{\shortstack{\textbf{GatedLinear}\\{\scriptsize (Ours)}}} & \multicolumn{2}{c|}{\shortstack{\textbf{DLinear}\\{\scriptsize (AAAI 2023)}}} & \multicolumn{2}{c|}{\shortstack{\textbf{TimeMixer}\\{\scriptsize (ICLR 2024)}}} & \multicolumn{2}{c|}{\shortstack{\textbf{MixLinear}\\{\scriptsize (ICLR 2026)}}} & \multicolumn{2}{c|}{\shortstack{\textbf{PaiFilter}\\{\scriptsize (NeurIPS 2024)}}} & \multicolumn{2}{c|}{\shortstack{\textbf{PatchTST}\\{\scriptsize (ICLR 2023)}}} & \multicolumn{2}{c|}{\shortstack{\textbf{PhaseFormer}\\{\scriptsize (ICLR 2026)}}} & \multicolumn{2}{c|}{\shortstack{\textbf{TQNet}\\{\scriptsize (ICML 2025)}}} & \multicolumn{2}{c|}{\shortstack{\textbf{TimeAlign}\\{\scriptsize (ICLR 2026)}}} & \multicolumn{2}{c}{\shortstack{\textbf{iTransformer}\\{\scriptsize (ICLR 2024)}}} \\
    \cmidrule(lr){2-3}\cmidrule(lr){4-5}\cmidrule(lr){6-7}\cmidrule(lr){8-9}\cmidrule(lr){10-11}\cmidrule(lr){12-13}\cmidrule(lr){14-15}\cmidrule(lr){16-17}\cmidrule(lr){18-19}\cmidrule(lr){20-21}
    & MSE & MAE & MSE & MAE & MSE & MAE & MSE & MAE & MSE & MAE & MSE & MAE & MSE & MAE & MSE & MAE & MSE & MAE & MSE & MAE \\
    \midrule
    \textbf{ECL} & \best{0.138} & \best{0.231} & 0.148 & 0.243 & 0.143 & \second{0.235} & 0.159 & 0.254 & \second{0.141} & 0.235 & 0.147 & 0.250 & 0.163 & 0.255 & 0.141 & 0.236 & 0.141 & 0.236 & 0.143 & 0.237 \\
    \textbf{ETTh1} & 0.404 & 0.414 & \second{0.388} & \second{0.403} & 0.400 & 0.415 & \best{0.383} & \best{0.399} & 0.433 & 0.440 & 0.401 & 0.417 & 0.400 & 0.407 & 0.391 & 0.409 & 0.402 & 0.415 & 0.459 & 0.452 \\
    \textbf{ETTh2} & \best{0.300} & \best{0.356} & \second{0.300} & \second{0.358} & 0.328 & 0.379 & 0.314 & 0.360 & 0.347 & 0.392 & 0.329 & 0.379 & 0.326 & 0.372 & 0.315 & 0.368 & 0.320 & 0.367 & 0.382 & 0.402 \\
    \textbf{ETTm1} & \best{0.305} & \best{0.344} & 0.322 & 0.356 & 0.325 & 0.366 & 0.338 & 0.371 & 0.330 & 0.369 & 0.317 & 0.362 & 0.540 & 0.484 & 0.313 & 0.355 & \second{0.307} & \second{0.350} & 0.357 & 0.385 \\
    \textbf{ETTm2} & \best{0.196} & \best{0.273} & \second{0.197} & \second{0.274} & 0.212 & 0.288 & 0.209 & 0.288 & 0.210 & 0.281 & 0.208 & 0.286 & 0.241 & 0.320 & 0.201 & 0.276 & 0.205 & 0.277 & 0.240 & 0.316 \\
    \textbf{Exchange} & \best{0.161} & \second{0.277} & 0.170 & \best{0.271} & 0.184 & 0.287 & \second{0.168} & 0.281 & 0.196 & 0.294 & 0.172 & 0.281 & 0.188 & 0.298 & 0.193 & 0.294 & 0.171 & 0.278 & 0.211 & 0.312 \\
    \textbf{Traffic} & 0.405 & 0.272 & 0.416 & 0.283 & 0.383 & 0.267 & 0.423 & 0.284 & 0.395 & 0.278 & 0.383 & 0.270 & 0.428 & 0.267 & \second{0.381} & \second{0.266} & 0.396 & 0.276 & \best{0.367} & \best{0.262} \\
    \textbf{Weather} & \best{0.173} & \best{0.218} & 0.198 & 0.240 & 0.178 & 0.225 & 0.207 & 0.252 & 0.179 & 0.226 & 0.179 & 0.221 & 0.208 & 0.257 & 0.182 & 0.226 & \second{0.174} & \second{0.218} & 0.191 & 0.235 \\
    \midrule
    \textbf{Average} & \best{0.260} & \best{0.298} & 0.267 & 0.303 & 0.269 & 0.308 & 0.275 & 0.311 & 0.279 & 0.314 & 0.267 & 0.308 & 0.312 & 0.333 & 0.265 & 0.304 & \second{0.264} & \second{0.302} & 0.294 & 0.325 \\
    \bottomrule
  \end{tabular}}
\end{table*}

Table~\ref{tab:main-results} summarizes the main forecasting results. Overall, GatedLinear obtains the best result on 11 out of 16 dataset-level metrics across the eight benchmarks. It also achieves the lowest average MSE and MAE, demonstrating the strongest overall forecasting performance among the compared methods. These results indicate that adaptively combining global, incremental, and phase-aligned linear bases can provide robust gains across heterogeneous time series datasets.

Although GatedLinear is not the best on every individual dataset, it remains competitive against recent linear, frequency-based, and Transformer-style baselines. The strong average performance suggests that the proposed adaptive mechanism-selection strategy offers a favorable balance between simplicity and accuracy, rather than relying on a single fixed forecasting mechanism.

\subsection{Ablation studies}

We conduct five groups of ablation experiments to examine the contribution of the forecasting bases and the structured fusion gate. Table~\ref{tab:ablation} reports the results on four representative datasets. The first three variants remove the global trend--seasonal branch, the difference-based incremental branch, and the phase-aligned recurrence branch, respectively, testing whether each basis provides useful complementary information. The fourth variant removes the phase-indexed logits from the fusion gate, which disables explicit future phase-aware routing. The fifth variant replaces the proposed Tri-Factorized Fusion Gate with a plain softmax gate over the three branches, whose weights are shared across horizons and variables.

\begin{table}[t]
  \caption{Ablation results on ECL, ETTm1, Traffic, and Weather. Numbers are averaged over prediction lengths $H\in\{48,96,192,336\}$ with input length $L=336$. Avg. Inc. reports the average relative increase (\%) over all eight MSE/MAE entries compared with the full model.}
  \label{tab:ablation}
  \centering
  \setlength{\tabcolsep}{3.5pt}
  \resizebox{0.8\linewidth}{!}{%
  \begin{tabular}{l|cc|cc|cc|cc|c}
    \toprule
    \multirow{2}{*}{\textbf{Variant}} & \multicolumn{2}{c|}{\textbf{ECL}} & \multicolumn{2}{c|}{\textbf{ETTm1}} & \multicolumn{2}{c|}{\textbf{Traffic}} & \multicolumn{2}{c|}{\textbf{Weather}} & \multirow{2}{*}{\shortstack{\textbf{Avg.}\\\textbf{Inc. (\%)}}} \\
    \cmidrule(lr){2-3}\cmidrule(lr){4-5}\cmidrule(lr){6-7}\cmidrule(lr){8-9}
    & MSE & MAE & MSE & MAE & MSE & MAE & MSE & MAE & \\
    \midrule
    \textbf{Full model} & \best{0.1378} & \best{0.2311} & \best{0.3054} & \best{0.3444} & \second{0.4046} & \best{0.2716} & \best{0.1726} & \best{0.2176} & \textbf{0.00} \\
    \textbf{w/o Global branch} & 0.1432 & 0.2372 & \second{0.3090} & \second{0.3457} & 0.4116 & 0.2774 & 0.1785 & 0.2202 & \textbf{+2.07} \\
    \textbf{w/o Difference branch} & \second{0.1389} & \second{0.2317} & 0.3108 & 0.3468 & 0.4144 & \second{0.2732} & 0.1823 & 0.2248 & \textbf{+1.93} \\
    \textbf{w/o Phase branch} & 0.1433 & 0.2377 & 0.3100 & 0.3468 & \best{0.4016} & 0.2742 & 0.1771 & 0.2196 & \textbf{+1.60} \\
    \textbf{w/o Phase-indexed logits} & 0.1423 & 0.2354 & 0.3112 & 0.3481 & 0.4158 & 0.2815 & \second{0.1752} & \second{0.2194} & \textbf{+2.11} \\
    \textbf{Plain softmax gate} & 0.1432 & 0.2368 & 0.3122 & 0.3492 & 0.4157 & 0.2819 & 0.1765 & 0.2205 & \textbf{+2.52} \\
    \bottomrule
  \end{tabular}}
\end{table}

The full model achieves the best overall performance across these ablations, indicating that the three bases and the structured gate jointly contribute to the final accuracy. Removing any single branch generally degrades performance, which suggests that global trend--seasonal projection, difference-based local evolution, and phase-aligned recurrence capture different temporal patterns and cannot be fully substituted by the remaining branches.

The gate-related ablations show a clearer performance drop, as reflected by the Avg. Inc. column. Removing the phase-indexed logits increases the averaged metric by 2.11\%, indicating that explicit future phase information is important for selecting among the branches. Replacing the Tri-Factorized Fusion Gate with a plain softmax gate leads to the largest increase of 2.52\%, because the same branch mixture must be shared across all horizons and variables. These results confirm that the improvement of GatedLinear comes not only from using multiple linear forecasting bases, but also from routing them with horizon-, channel-, and phase-aware information.

\subsection{Interpretability analysis}

\begin{figure}[htbp]
  \centering
  \includegraphics[width=\linewidth]{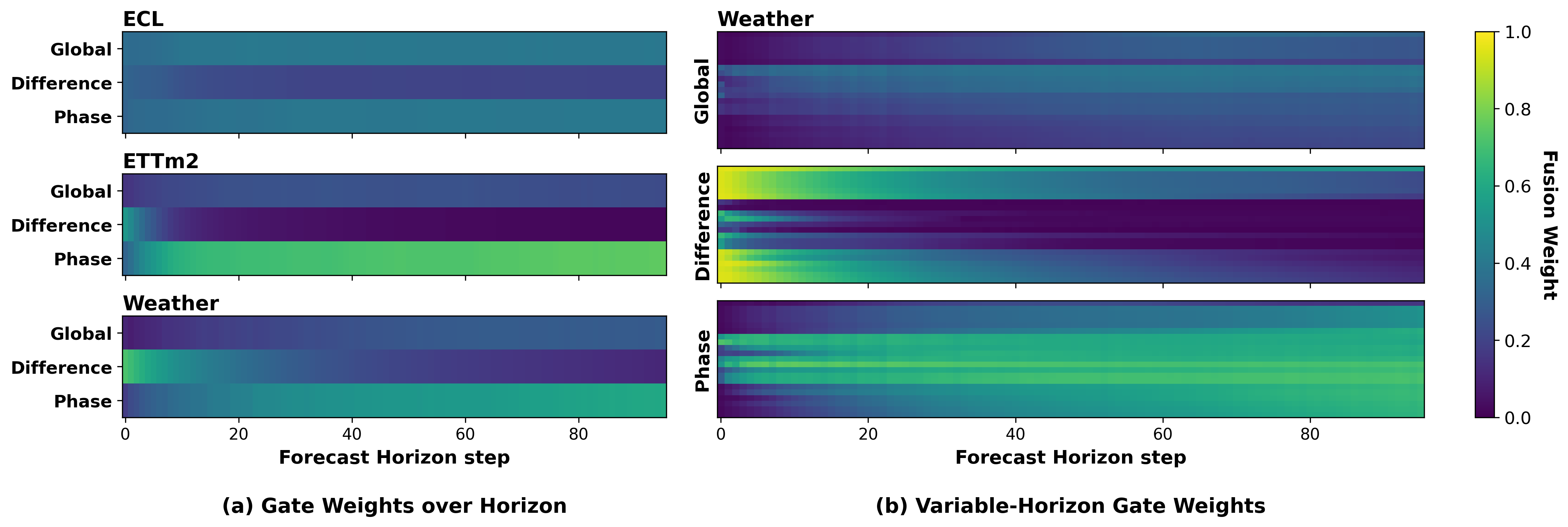}
  \caption{Learned fusion gate weights across horizons and variables. (a) Different datasets allocate branch weights over the forecast horizon. (b) Variable- and horizon-specific gate weights on the Weather dataset.}
  \label{fig:fusion-gate-horizon-variable}
\end{figure}

The structured gate provides a direct diagnostic view of mechanism selection. Figure~\ref{fig:fusion-gate-horizon-variable}(a) shows that different datasets learn different branch-weight preferences over the forecast horizon, indicating that the model does not rely on a fixed mixture of forecasting mechanisms. Instead, it adjusts the relative importance of the global, difference, and phase-aligned branches according to dataset-specific temporal properties. A consistent pattern is that the difference branch tends to receive larger weights at shorter horizons and gradually decreases as the horizon extends, while the phase-aligned branch often becomes more prominent at longer horizons. This behavior is intuitive: difference-based incremental prediction is useful for capturing near-term local evolution, but its errors may accumulate over longer forecasts; in contrast, the phase-aligned branch can provide a more stable reference by reusing recurring periodic structure. The gate therefore dynamically shifts the mixture from short-term incremental evolution toward more reliable long-horizon periodic cues when appropriate.

Figure~\ref{fig:fusion-gate-horizon-variable}(b) further shows that even within the same dataset, different variables can have distinct branch preferences. In each branch-specific rectangular heatmap, each row corresponds to one variable channel and each column corresponds to a forecast horizon step. The heterogeneous row patterns suggest that GatedLinear learns variable-specific routing behavior, allowing each variable to adaptively select the branch mixture that matches its own temporal characteristics.

\begin{figure}[htbp]
  \centering
  \includegraphics[width=\linewidth]{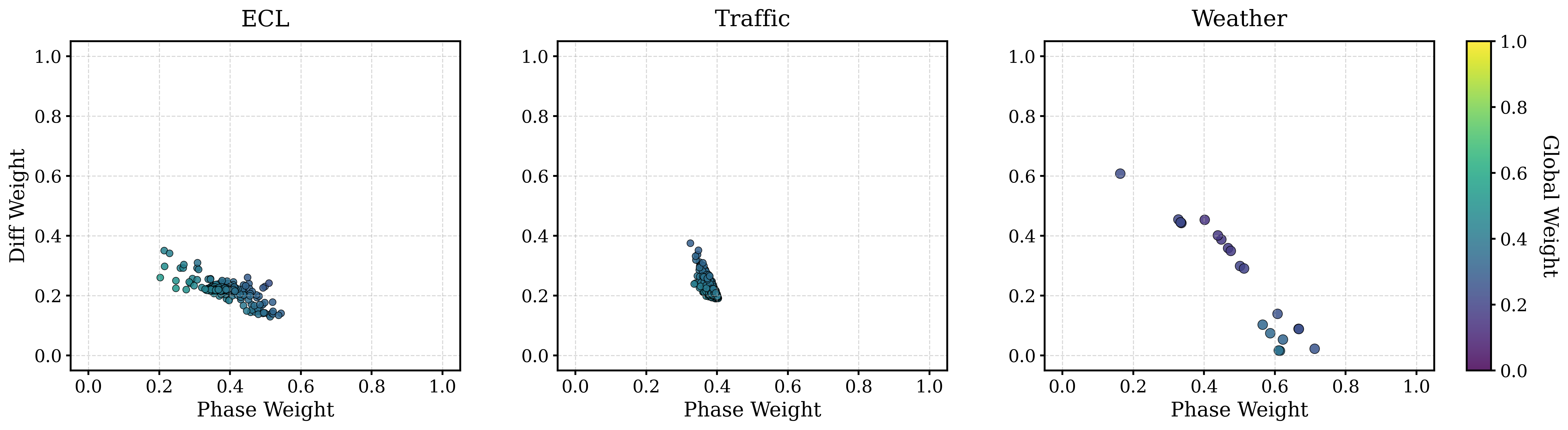}
  \caption{Channel-wise branch-weight preferences across different datasets. Each point denotes a variable channel.}
  \label{fig:channel-preference-scatter}
\end{figure}

Figure~\ref{fig:channel-preference-scatter} visualizes channel-wise branch preferences for ECL, Traffic, and Weather. Each point denotes one variable, and its position reflects the relative reliance on the phase-aligned and difference-based branches. The Traffic variables form a compact cluster, while the Weather variables are much more dispersed. This pattern is consistent with the channel-level trait distributions in Figure~\ref{fig:intro-traits}(b): Weather exhibits stronger variable-level heterogeneity, whereas Traffic shows smaller variation among variables. Together, Figures~\ref{fig:fusion-gate-horizon-variable} and~\ref{fig:channel-preference-scatter} show that the learned gate preferences are aligned with the underlying data characteristics at both the dataset and variable levels.

To further quantify this alignment, Figure~\ref{fig:channel-spearman-heatmap} aggregates all variable channels from the evaluated datasets and reports the Spearman correlations between channel-level temporal traits and learned branch preferences. The difference branch is closely associated with shift-related traits such as Mean Shift and Scale Shift, indicating that it is emphasized for variables that require incremental evolution. In contrast, the phase branch is positively correlated with Seasonality Strength and Phase Consistency, matching its intended role of exploiting stable recurring phases. These correlations show that the gate learns semantically meaningful branch preferences rather than arbitrary weights, further supporting the interpretability and effectiveness of the adaptive routing mechanism.

\subsection{Efficiency analysis}

\begin{figure}[htbp]
  \centering
  \begin{minipage}[t]{0.49\linewidth}
    \centering
    \includegraphics[width=\linewidth]{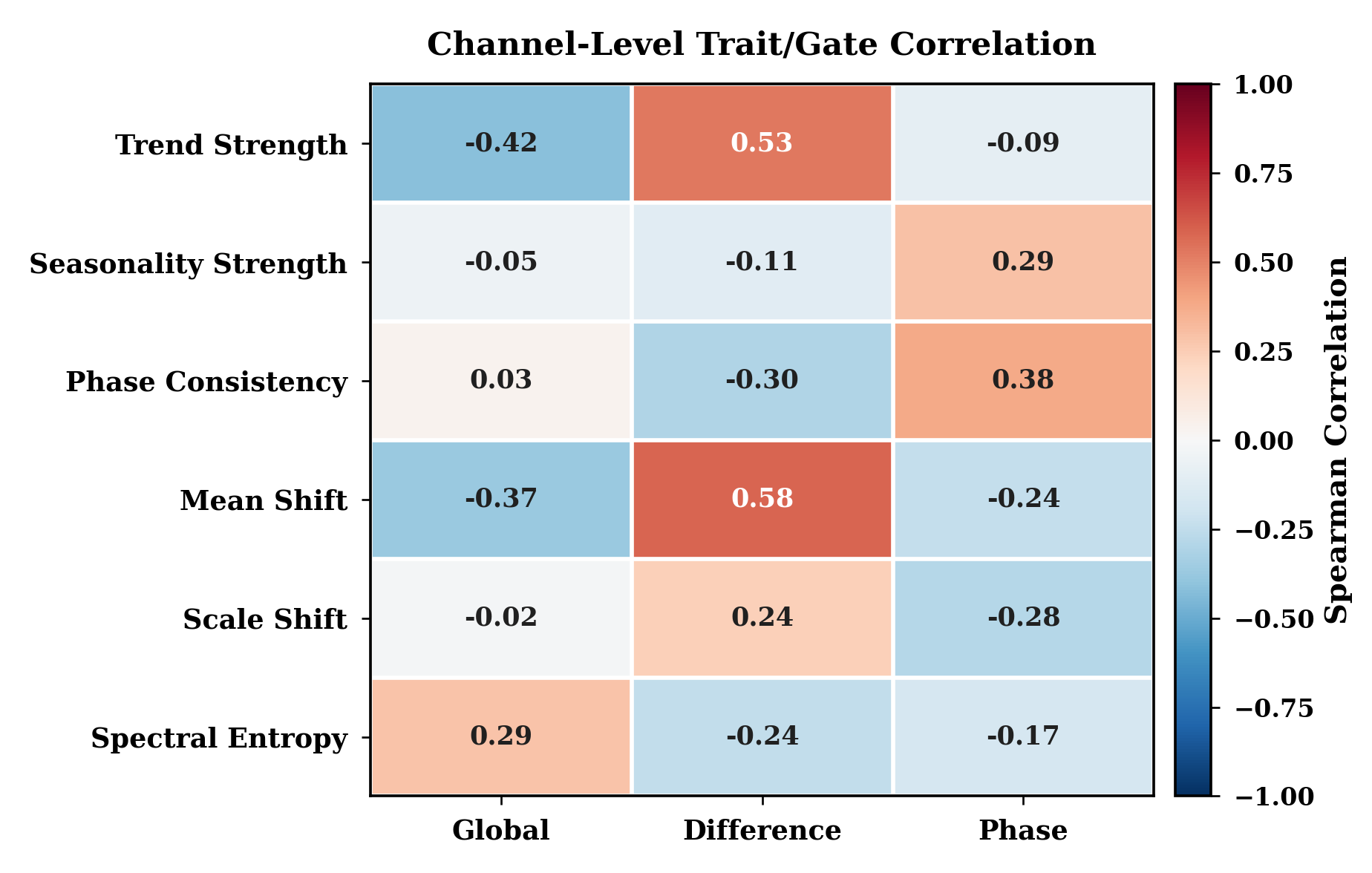}
    \caption{Spearman correlation heatmap between channel-level temporal traits and learned branch preferences. The computation of Spearman correlations is provided in Appendix~\ref{app:trait-gate-correlation}.}
    \label{fig:channel-spearman-heatmap}
  \end{minipage}
  \hfill
  \begin{minipage}[t]{0.49\linewidth}
    \centering
    \includegraphics[width=\linewidth]{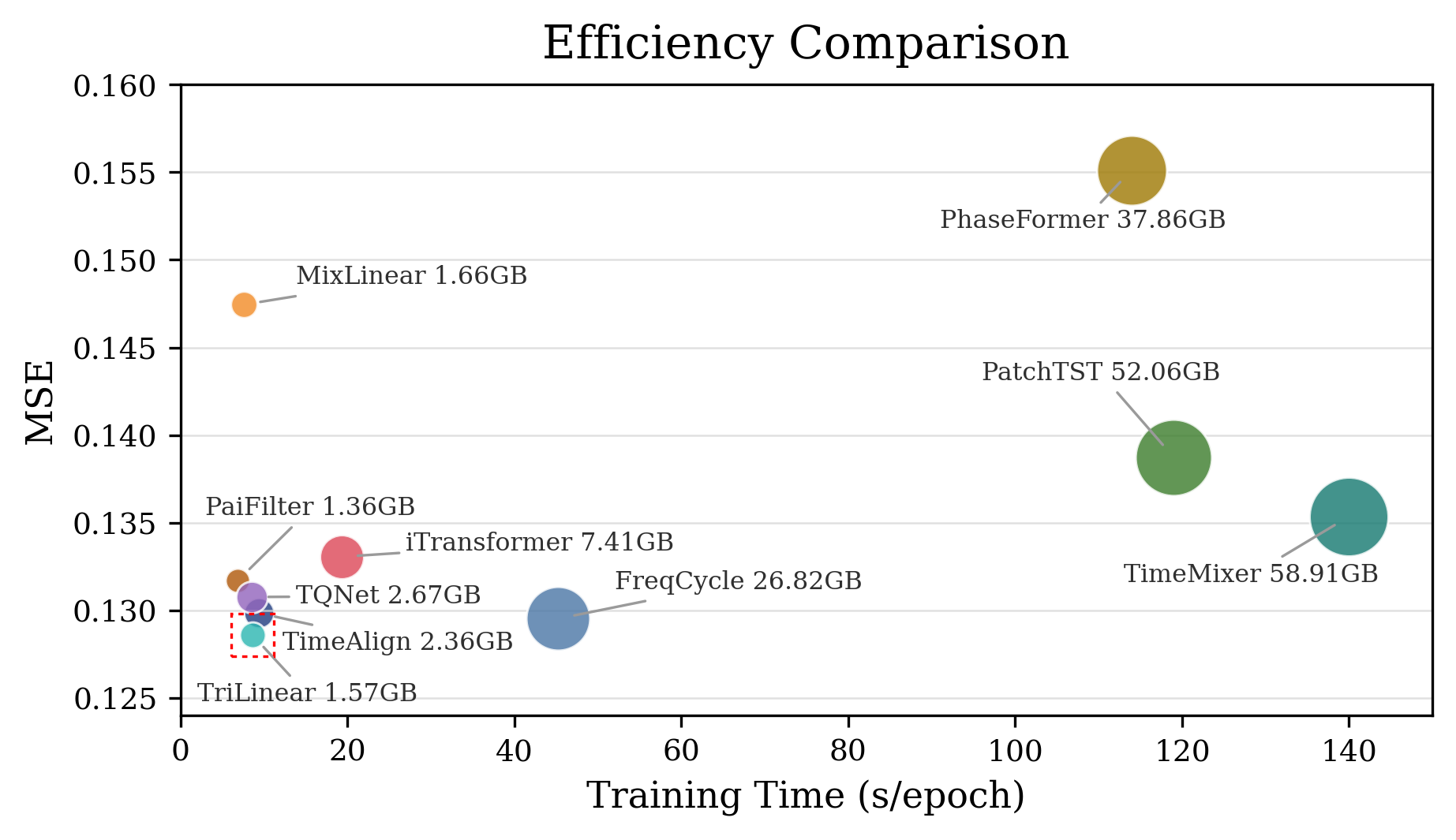}
    \caption{Efficiency comparison on the Electricity dataset using input length $L=336$, prediction horizon $H=96$, and batch size 128. Bubble size indicates peak GPU memory.}
    \label{fig:efficiency-ecl}
  \end{minipage}
\end{figure}

Figure~\ref{fig:efficiency-ecl} compares forecasting accuracy and computational efficiency on the Electricity dataset. The proposed model lies in the favorable lower-left region, showing low forecasting error with short training time and small memory consumption. This efficiency comes from using only three lightweight forecasting bases together with a compact structured gate, avoiding the expensive attention or deep mixing modules used by heavier baselines. Overall, the results show that GatedLinear achieves a strong accuracy-efficiency trade-off while preserving a small computational footprint.

\section{Conclusion}

We introduced GatedLinear, a lightweight forecasting framework built on the principle that heterogeneous time series should not be forced through a single fixed predictive mechanism. Instead of increasing backbone complexity, GatedLinear formulates forecasting as adaptive routing over three complementary linear bases: a global trend--seasonal projection for smooth extrapolation, a difference-based incremental basis for local nonstationary evolution, and a phase-aligned recurrence basis for explicit cyclic reuse. This tri-basis design provides a compact yet expressive mechanism pool whose components correspond to distinct temporal regimes.

The key innovation of GatedLinear is the Tri-Factorized Fusion Gate, which performs forecast-point-level soft selection by factorizing routing logits into channel-specific preferences, horizon-dependent offsets, and future phase-indexed biases. This structure allows the model to adapt across variables, prediction steps, and periodic phases while keeping the predictor efficient and interpretable. Experiments on standard benchmarks show that GatedLinear achieves the best average MSE and MAE and remains highly competitive with recent linear, frequency-based, and Transformer-style forecasters. Ablation and gate analyses further verify that the gains arise from both the complementary forecasting bases and the semantically meaningful routing among them: the model emphasizes incremental evolution under shift-like dynamics and phase recurrence under stable periodicity. These results suggest that carefully structured mechanism selection can be a powerful alternative to stacking heavier neural modules for time series forecasting.

\clearpage

\bibliographystyle{unsrtnat}
\bibliography{references}


\appendix

\section{Additional Related Work}
\label{app:related-work}

\paragraph{Efficient linear and MLP-based forecasting.}
Long-sequence forecasting has recently seen a clear shift from increasingly complex attention backbones toward Transformer-free architectures that rely on simple temporal mappings, fully connected blocks, and MLP-style mixing. N-BEATS~\citep{oreshkin2019n} introduced a deep fully connected architecture based on residual stacks and neural basis expansion, while N-HiTS~\citep{challu2023nhits} improved this basis-style view through hierarchical interpolation and multi-rate sampling for long-horizon forecasting. DLinear~\citep{zeng2023transformers} demonstrates that decomposing a series into trend and seasonal components followed by one-layer linear mappings can already form a highly competitive baseline. TSMixer~\citep{chen2023tsmixer} uses all-MLP time- and feature-mixing blocks to capture temporal and cross-variate information, TimeMixer~\citep{wangtimemixer} extends this efficiency-oriented view with decomposable multi-scale MLP mixing, and MixLinear~\citep{ma2024mixlinear} further targets low-resource multivariate forecasting through segment-based temporal modeling and adaptive low-rank spectral filtering. These methods show that lightweight predictors can challenge heavier Transformer-style models, but they usually instantiate one dominant forecasting mechanism for all variables and forecast positions. GatedLinear follows this efficient forecasting lineage, yet treats lightweight linear operations as complementary bases rather than as a single fixed backbone, allowing smooth projection, incremental evolution, and periodic recurrence to be selected adaptively.

\paragraph{Periodicity and phase-aware modeling.}
Another line of work explicitly exploits frequency structure, cyclic recurrence, or phase alignment in time series. FEDformer~\citep{zhou2022fedformer} introduces frequency-enhanced decomposition inside a Transformer architecture, TimesNet~\citep{wutimesnet} discovers dominant periods and converts one-dimensional sequences into two-dimensional period-aligned representations, and FilterNet~\citep{yi2024filternet} learns frequency-domain filters for shaping informative spectral components. CycleNet~\citep{lin2024cyclenet} models periodic patterns with learnable recurrent cycles and predicts residual components, while FreqCycle~\citep{zhang2026freqcycle} combines shared-cycle learning in the time domain with segmented frequency modeling for residual components. PhaseFormer~\citep{niu2025phaseformer} replaces patch tokens with phase-wise modeling and lightweight cross-phase routing for periodic forecasting. These approaches encode periodicity through spectral transforms, learned filters, learnable cycles, period-to-2D transformations, or phase-aware tokenization. In contrast, GatedLinear uses a direct time-domain recurrence basis that reuses historical values at matched future phases, and its phase-indexed gate logits decide when such exact phase reuse should dominate over the other bases.

\paragraph{Dataset heterogeneity and adaptive routing.}
Comprehensive benchmark studies suggest that forecasting data are highly heterogeneous across domains, variables, and temporal regimes. TFB~\citep{qiu2024tfb} broadens evaluation coverage across diverse real-world domains and characterizes datasets to support fairer comparisons among statistical, machine-learning, and deep-learning forecasters. SynTSBench~\citep{tansyntsbench} further constructs synthetic datasets with programmable temporal patterns and shows that existing deep forecasting models rarely perform uniformly well across all pattern types. These observations support the view that trend strength, periodicity, stationarity, and local drift vary substantially across datasets and even across variables within the same dataset. Adaptive routing is therefore a natural response: mixture-of-experts models~\citep{jacobs1991adaptive,shazeer2017outrageously} use gates to assign inputs to specialized predictors, and recent time-series architectures such as Pathformer~\citep{chen2024pathformer} and PhaseFormer~\citep{niu2025phaseformer} introduce adaptive pathways or routers for scale and phase interactions. GatedLinear shares this motivation but routes among interpretable linear bases instead of deep experts, with routing logits factorized by channel, horizon, and future phase so that each forecast point can select the mechanism most compatible with its temporal traits.

\section{Limitations of Our Work}
\label{app:limitations}

\paragraph{Channel-independent design.}
GatedLinear follows the common channel-independent forecasting setting: the temporal linear branches are shared across variables and each channel is mainly forecast from its own historical trajectory. Although the fusion gate learns channel-specific routing preferences, the current design does not explicitly model cross-variable dependencies. This choice keeps the model lightweight and interpretable, but it may be less effective when multivariate interactions are central to the forecasting problem. A representative case is Traffic, which contains 862 sensor variables; on this dataset, methods that directly model variate-level relations, such as iTransformer and TQNet, obtain stronger results.

\paragraph{Dependence on meaningful periods and time marks.}
The phase-aligned recurrence basis and the phase-indexed gate assume that the data contain a usable base period and that future phase indices are meaningful. When future time marks are available, these indices can be obtained from the decoder-side timestamps; otherwise, the model falls back to horizon-based phase indices. This assumption may be weakened for weakly periodic series, drifting periods, holiday or event effects, abrupt disruptions, and irregularly sampled observations. In such cases, reusing historical values from the same phase may become less reliable.

\paragraph{Limited expressiveness of the linear basis pool.}
Our framework intentionally restricts its forecasting mechanisms to three simple linear bases: smooth trend--seasonal projection, difference-based local evolution, and phase-aligned recurrence with a linear residual. This restriction is important for the model's efficiency and interpretability, but it also bounds the expressive power of the predictor. Highly nonlinear dynamics, abrupt regime shifts, or trajectories driven by external events may not be fully captured by these three branches alone.

\section{Theoretical Motivation of the Forecasting Bases}
\label{app:theoretical-analysis}

To formalize the intuition behind our multi-basis design, we provide a theoretical analysis illustrating why a single global forecasting mechanism struggles with non-stationary data, and why incorporating the difference-based incremental basis is theoretically optimal in such scenarios.

\subsection{Optimality of the Difference-Based Basis on Non-Stationary Data}

Consider a data generating process governed by an integrated random walk with drift (e.g., an ARIMA(0,1,0) process), which is a standard mathematical model for non-stationary series with level shifts. The process is defined as:
\begin{equation}
  x_t = x_{t-1} + c + \epsilon_t, \quad \epsilon_t \sim \mathcal{N}(0, \sigma^2),
\end{equation}
where $c$ is a constant deterministic drift and $\epsilon_t$ is independent zero-mean Gaussian noise. Unrolling this process gives the true future value at target step $T+h$:
\begin{equation}
  x_{T+h} = x_T + h c + \sum_{i=1}^{h} \epsilon_{T+i}.
\end{equation}
The optimal Minimum Mean Squared Error (MMSE) predictor is $\mathbb{E}[x_{T+h} | x_{\le T}] = x_T + h c$, with a theoretical minimum error variance of $h\sigma^2$.

\paragraph{Suboptimality of the Global Basis.}
A standard global linear mapping (or similarly, global self-attention over raw values) attempts to forecast the future by computing a weighted combination of historical observations: $\hat{x}^{(global)}_{T+h} = \sum_{i=1}^{L} w_{h,i} x_{T-L+i}$.
If the test data experiences a structural \textit{level shift} (i.e., the absolute magnitude of $x$ is shifted by an arbitrary scalar $\Delta_{\mathrm{shift}}$ not seen in the training set), the prediction becomes:
\begin{equation}
  \hat{x}^{(global)}_{T+h}(\text{shifted}) = \sum_{i=1}^{L} w_{h,i} (x_{T-L+i} + \Delta_{\mathrm{shift}}) = \hat{x}^{(global)}_{T+h} + \Delta_{\mathrm{shift}} \sum_{i=1}^{L} w_{h,i}.
\end{equation}
To maintain shift-invariance (i.e., for the shift in input to properly propagate to a shift in prediction without scaling error), the learned weights must strictly satisfy $\sum w_{h,i} = 1$. In practice, achieving and perfectly maintaining this strict equality constraint via unconstrained stochastic gradient descent is highly improbable, especially under weight decay regularization. Consequently, any level shift in the input induces an unconditional bias $\mathcal{O}(\Delta_{\mathrm{shift}})$, causing the generalization error variance to grow unbounded with the shift magnitude.

\paragraph{Optimality of the Difference-Based Basis.}
Our difference-based incremental branch mathematically perfectly resolves this issue. By computing the first-order difference $\Delta x_t = x_t - x_{t-1} = c + \epsilon_t$, the input sequences become strictly stationary and entirely independent of the absolute level of $x_t$. The linear layer $\mathcal{L}_d$ now maps from a stationary sequence drawn from $\mathcal{N}(c, \sigma^2)$ to smoothly predict the stationary target increment $c$:
\begin{equation}
  \Delta\hat{x}_{T+i} = c.
\end{equation}
Crucially, the future values are reconstructed by integrating these predicted increments starting strictly from the last historically observed value $x_T$:
\begin{equation}
  \hat{x}^{(diff)}_{T+h} = x_T + \sum_{i=1}^{h} \Delta\hat{x}_{T+i} = x_T + h c.
\end{equation}
This formulation is structurally identical to the theoretical MMSE predictor. The estimation of the required dynamic increments is completely decoupled from the absolute scale of the sequence, rendering the basis strictly invariant to arbitrary level shifts $\Delta_{\mathrm{shift}}$. Its theoretical error variance is bounded exactly to the system's inherent noise variance $h\sigma^2$. This demonstrates that integrating the difference-based recurrence is not merely an empirical engineering decision, but a theoretically necessary inductive bias to guarantee robust and unbounded-error-free forecasting under non-stationary drifting environments.

\section{Temporal Trait and Gate-Correlation Analysis}
\label{app:trait-gate-correlation}

This section provides the definitions used to construct the temporal-trait visualization in Figure~\ref{fig:intro-traits} and the channel-level gate-correlation heatmap in Figure~\ref{fig:channel-spearman-heatmap}. For a single variable channel, let the training sequence be $x_{1:T}=(x_1,\ldots,x_T)$, where $T$ is the training length. Let $P$ denote the dataset-specific base period, e.g., $P=24$ for hourly ETT data and $P=168$ for weekly hourly datasets such as Electricity and Traffic. For multivariate datasets, all traits are computed independently for each channel.

\paragraph{\textbf{Trend and seasonality strength}.}
We first estimate a low-frequency trend component using a centered moving average with window size $P$,
\begin{equation}
  \tau_t = \mathrm{MA}_P(x)_t .
\end{equation}
The detrended sequence is $d_t=x_t-\tau_t$. We then estimate the seasonal component by averaging detrended values that share the same phase,
\begin{equation}
  s_t =
  \frac{1}{|\mathcal{I}_{\phi(t)}|}
  \sum_{i\in\mathcal{I}_{\phi(t)}} d_i,
  \qquad
  \phi(t)=t\bmod P,
\end{equation}
where $\mathcal{I}_{\phi}=\{i:i\bmod P=\phi\}$. The residual is $r_t=x_t-\tau_t-s_t$. Trend strength and seasonality strength are then computed as
\begin{equation}
  \mathrm{TrendStrength}
  =
  1-
  \frac{\mathrm{Var}(r)}
  {\max(\mathrm{Var}(x-s),\epsilon)},
  \qquad
  \mathrm{SeasonalityStrength}
  =
  1-
  \frac{\mathrm{Var}(r)}
  {\max(\mathrm{Var}(x-\tau),\epsilon)} ,
\end{equation}
and clipped to $[0,1]$. Here $\epsilon$ is a small constant for numerical stability.

\paragraph{\textbf{Phase consistency}.}
Phase consistency measures whether values at the same phase of the base period are reproducible across cycles. Let $K=\lfloor T/P\rfloor$ be the number of complete cycles and fold the sequence into $X_{k,p}$ for $k=1,\ldots,K$ and $p=0,\ldots,P-1$. Denote
\begin{equation}
  \bar{X}_{p}=\frac{1}{K}\sum_{k=1}^{K}X_{k,p},
  \qquad
  \bar{X}=\frac{1}{KP}\sum_{k=1}^{K}\sum_{p=0}^{P-1}X_{k,p}.
\end{equation}
The phase-consistency score is the fraction of total variance explained by phase-specific means:
\begin{equation}
  \mathrm{PhaseConsistency}
  =
  \frac{
    K\sum_{p=0}^{P-1}(\bar{X}_{p}-\bar{X})^2
  }{
    \sum_{k=1}^{K}\sum_{p=0}^{P-1}(X_{k,p}-\bar{X})^2+\epsilon
  } .
\end{equation}
A larger value indicates more stable phase-aligned recurrence and therefore stronger support for the phase-aligned branch.

\paragraph{\textbf{Mean and scale shift}.}
Mean shift captures low-frequency level nonstationarity. For a rolling window of length $P$, define
\begin{equation}
  \mu_t^{(P)} = \frac{1}{P}\sum_{i=t}^{t+P-1}x_i .
\end{equation}
The mean-shift score is
\begin{equation}
  \mathrm{MeanShift}
  =
  \mathrm{Std}\left(\{\mu_t^{(P)}\}_{t=1}^{T-P+1}\right).
\end{equation}
Scale shift captures time-varying volatility or amplitude. With rolling standard deviation
\begin{equation}
  \sigma_t^{(P)}
  =
  \sqrt{
    \frac{1}{P}\sum_{i=t}^{t+P-1}
    \left(x_i-\mu_t^{(P)}\right)^2
  },
\end{equation}
we compute
\begin{equation}
  \mathrm{ScaleShift}
  =
  \frac{
    \mathrm{Std}\left(\{\sigma_t^{(P)}\}_{t=1}^{T-P+1}\right)
  }{
    \mathrm{Mean}\left(\{\sigma_t^{(P)}\}_{t=1}^{T-P+1}\right)+\epsilon
  } .
\end{equation}
Larger values indicate stronger level drift or scale nonstationarity.

\paragraph{\textbf{Spectral entropy}.}
Spectral entropy measures how concentrated the frequency-domain energy is. Let $S(f_j)$ be the power spectral density estimated by Welch's method and normalize it as
\begin{equation}
  p_j=\frac{S(f_j)}{\sum_m S(f_m)} .
\end{equation}
The normalized spectral entropy is
\begin{equation}
  \mathrm{SpectralEntropy}
  =
  -\frac{1}{\log_2 M}
  \sum_{j=1}^{M}p_j\log_2(p_j+\epsilon),
\end{equation}
where $M$ is the number of frequency bins. A larger value indicates more dispersed spectral energy and more complex frequency composition.

\paragraph{\textbf{Robust normalization for visualization}.}
For the radar and boxplot visualizations, each trait is robust-normalized across all selected channel samples. Given channel-level trait values $\{z_i\}_{i=1}^{N}$, let $q_2$ and $q_{98}$ be the 2nd and 98th percentiles. The normalized score is
\begin{equation}
  \tilde{z}_i
  =
  \mathrm{clip}
  \left(
  \frac{z_i-q_2}{q_{98}-q_2+\epsilon},
  0,1
  \right).
\end{equation}
Dataset-level radar values are obtained by averaging these robust-normalized channel-level scores.

\paragraph{\textbf{Gate weight aggregation and correlation}.}
The fusion gate produces branch weights $w_{h,c,b}$ for forecast horizon $h$, channel $c$, and branch $b\in\{\mathrm{Global},\mathrm{Difference},\mathrm{Phase}\}$. For channel-level correlation analysis, we aggregate each channel's branch weight over the prediction horizon:
\begin{equation}
  \bar{w}_{c,b}
  =
  \frac{1}{H}\sum_{h=1}^{H} w_{h,c,b}.
\end{equation}
The heatmap in Figure~\ref{fig:channel-spearman-heatmap} reports Spearman rank correlation between channel-level temporal traits and aggregated branch preferences over all selected channel samples. We use Spearman correlation because it captures monotonic relationships and is less sensitive to outliers or nonlinear scaling than Pearson correlation.

For completeness, Pearson correlation between trait values $a_i$ and branch weights $b_i$ is
\begin{equation}
  \rho_{\mathrm{Pearson}}(a,b)
  =
  \frac{
    \sum_{i=1}^{N}(a_i-\bar{a})(b_i-\bar{b})
  }{
    \sqrt{\sum_{i=1}^{N}(a_i-\bar{a})^2}
    \sqrt{\sum_{i=1}^{N}(b_i-\bar{b})^2}
  } .
\end{equation}
Spearman correlation first converts the two variables into ranks and then computes Pearson correlation on the ranked values:
\begin{equation}
  \rho_{\mathrm{Spearman}}(a,b)
  =
  \rho_{\mathrm{Pearson}}\left(R(a),R(b)\right).
\end{equation}

\section{Experimental Details}
\label{app:experimental-details}

\subsection{Datasets}
\label{app:datasets}

We provide additional details for the forecasting benchmarks used in our experiments.
\begin{itemize}
  \item The ETT benchmark contains two sub-datasets: ETTh, collected at 1-hour intervals, and ETTm, collected at 15-minute intervals. Both were recorded from electricity transformers between July 2016 and July 2018 \citep{zhou2021informer}.
  \item Electricity records the hourly electricity consumption of 321 clients from 2012 to 2014 \citep{wu2021autoformer}.
  \item Traffic provides hourly measurements from 862 sensors on San Francisco Bay Area freeways from January 2015 to December 2016 \citep{wu2021autoformer}.
  \item Weather includes 21 meteorological indicators recorded every 10 minutes during 2020 at the Weather Station of the Max Planck Biogeochemistry Institute \citep{wu2021autoformer}.
  \item Exchange contains daily exchange rates for eight countries, including Australia, Britain, Canada, Switzerland, China, Japan, New Zealand, and Singapore, from 1990 to 2016 \citep{lai2018modeling}.
\end{itemize}
Unless otherwise stated, each dataset is split into train/validation/test subsets with a ratio of 7:1:2. The detailed dataset statistics are summarized in Table~\ref{tab:dataset-statistics}.

\begin{table}[h]
  \caption{Dataset statistics.}
  \label{tab:dataset-statistics}
  \centering
  \begin{tabular}{lcccc}
    \toprule
    Dataset & Frequency & Duration & Features & Samples \\
    \midrule
    ETTh & 1 hour & 2016--2018 & 7 & 17,420 \\
    ETTm & 15 min & 2016--2018 & 7 & 69,680 \\
    Electricity & 1 hour & 2012--2014 & 321 & 26,304 \\
    Traffic & 1 hour & 2015--2016 & 862 & 17,544 \\
    Weather & 10 min & 2020 & 21 & 52,696 \\
    Exchange & 1 day & 1990--2016 & 8 & 7,588 \\
    \bottomrule
  \end{tabular}
\end{table}

The base period $P$ is set according to the dataset frequency: $P=24$ for ETTh1/ETTh2, $P=96$ for ETTm1/ETTm2, $P=168$ for Electricity and Traffic, $P=144$ for Weather, and $P=24$ for Exchange.

\subsection{Evaluation Protocol and Metrics}
\label{app:evaluation-protocol}

All real-dataset experiments use input length $L=336$, prediction lengths $H\in\{48,96,192,336\}$, and label length 0. We report mean squared error (MSE) and mean absolute error (MAE), following standard practice in time series forecasting. Lower values indicate better performance. Reported numbers are averaged over three random seeds, and the validation set is used for early stopping and hyperparameter selection.

\subsection{Baselines}
\label{app:baselines}

We compare GatedLinear with the representative baselines reported in Table~\ref{tab:main-results}. Their model categories and source repositories are summarized below.

\paragraph{\textbf{DLinear}.}
DLinear is a simple linear forecasting model that decomposes each input sequence into trend and seasonal components and applies one-layer linear mappings to both parts. Despite its simplicity, it provides a strong and efficient baseline for long-term time series forecasting. The official implementation is available at \url{https://github.com/cure-lab/LTSF-Linear}.

\paragraph{\textbf{TimeMixer}.}
TimeMixer is an MLP-based forecasting model that exploits decomposable multi-scale patterns. It mixes seasonal and trend information across multiple temporal resolutions and aggregates multi-scale predictors for future forecasting. The official implementation is available at \url{https://github.com/kwuking/TimeMixer}.

\paragraph{\textbf{MixLinear}.}
MixLinear is an ultra-lightweight multivariate forecasting model designed for resource-constrained settings. It combines segment-based temporal modeling with adaptive low-rank spectral filtering, reducing the parameter scale while preserving both temporal and frequency-domain information. The official implementation is available at \url{https://github.com/aitianma/MixLinear}.

\paragraph{\textbf{PaiFilter}.}
PaiFilter is a plain shaping filter introduced in FilterNet. It learns a universal frequency-domain filter to selectively shape time series signals and is recommended by FilterNet for datasets with a relatively small number of variables. The official implementation is available at \url{https://github.com/aikunyi/FilterNet}.

\paragraph{\textbf{PatchTST}.}
PatchTST adapts Transformers to long-term time series forecasting by segmenting each series into patch tokens and using channel-independent processing. This design enables the model to capture long-range temporal dependencies with improved efficiency. The official implementation is available at \url{https://github.com/yuqinie98/PatchTST}.

\paragraph{\textbf{PhaseFormer}.}
PhaseFormer replaces patch-based tokenization with phase-wise modeling for periodic time series. It uses compact phase embeddings and a lightweight cross-phase routing mechanism to capture periodic dynamics with very low parameter cost. The official implementation is available at \url{https://github.com/neumyor/PhaseFormer_TSL}.

\paragraph{\textbf{TQNet}.}
TQNet introduces temporal query vectors to model multivariate correlations efficiently. It uses periodically shifted learnable vectors as queries in an attention mechanism, combining global temporal priors with sample-specific key-value representations. The official implementation is available at \url{https://github.com/ACAT-SCUT/TQNet}.

\paragraph{\textbf{TimeAlign}.}
TimeAlign is a lightweight plug-and-play alignment framework for time series forecasting. It aligns past and future representations through a reconstruction branch and a distribution-aware alignment module, helping reduce the mismatch between historical inputs and future targets. The official implementation is available at \url{https://github.com/TROUBADOUR000/TimeAlign}.

\paragraph{\textbf{iTransformer}.}
iTransformer inverts the conventional Transformer design by treating variates as tokens and modeling multivariate correlations through attention. This architecture improves the effectiveness of Transformer modules on multivariate time series forecasting tasks. The official implementation is available at \url{https://github.com/thuml/iTransformer}.

\subsection{Implementation Details}
\label{app:implementation-details}

The code implementation uses four linear branches: seasonal, trend, difference, and phase residual. The seasonal and trend branches jointly form the decomposed direct basis. The difference branch predicts increments and reconstructs future levels with cumulative summation. The phase branch averages the most recent complete cycles into a phase template and adds a learned residual. Fusion logits are stored as a base table, a horizon table, and a phase table. When available, future time marks select phase logits; otherwise, horizon modulo period is used.

Our model is trained for at most 30 epochs using Adam with mean squared error loss, an initial learning rate of 0.005, batch size 128, and early stopping patience 20 based on validation loss. The moving-average kernel for trend--seasonal decomposition is set to 25. For the Tri-Factorized Fusion Gate, we use fusion temperature $\tau=0.8$. RevIN is enabled with $\epsilon=10^{-5}$ and without affine parameters.

The phase branch uses the most recent available complete cycles, with a target of 3 cycles for ETTh1 and 2 cycles for the other datasets, clipped by the number of complete cycles contained in the input window. Future time marks are used to select phase-indexed gate logits when the phase can be decoded; otherwise, the model falls back to horizon-modulo-period indexing.

\section{Additional Empirical Results}
\label{app:additional-results}

\subsection{Detailed Forecasting Results}
\label{app:detailed-results}

Table~\ref{tab:detailed-results} reports the detailed MSE and MAE results for input length $L=336$ and prediction lengths $H\in\{48,96,192,336\}$.
The horizon-wise breakdown further clarifies the averaged results in Table~\ref{tab:main-results}. GatedLinear is consistently strong on ECL, ETTm1, ETTm2, ETTh2, Exchange, and Weather, where it achieves the best or second-best results for most horizons. The gains are especially clear on ECL and Weather as the prediction horizon becomes longer, suggesting that the adaptive combination of global, difference-based, and phase-aligned bases helps maintain stable forecasts beyond the short-horizon regime. On ETTh1 and Traffic, specialized baselines remain stronger for some horizons, indicating that a fixed lightweight channel-independent design is not uniformly optimal for every dataset. Nevertheless, GatedLinear remains competitive across these cases and obtains the best overall average performance, supporting the claim that adaptive mechanism selection provides a robust accuracy profile across heterogeneous benchmarks.

\begin{table}[p]
  \caption{Detailed forecasting results with input length $L=336$. The best results are shown in \best{red bold}, and the second-best in \second{blue underline}. Lower MSE/MAE is better.}
  \label{tab:detailed-results}
  \centering
  \scriptsize
  \setlength{\tabcolsep}{2.2pt}
  \renewcommand{\arraystretch}{1.22}
  \resizebox{\linewidth}{!}{%
  \begin{tabular}{ll|cc|cc|cc|cc|cc|cc|cc|cc|cc|cc}
    \toprule
    \multirow{2}{*}{\textbf{Dataset}} & \multirow{2}{*}{\textbf{$H$}} & \multicolumn{2}{c}{\textbf{GatedLinear}} & \multicolumn{2}{c}{\textbf{DLinear}} & \multicolumn{2}{c}{\textbf{TimeMixer}} & \multicolumn{2}{c}{\textbf{MixLinear}} & \multicolumn{2}{c}{\textbf{PaiFilter}} & \multicolumn{2}{c}{\textbf{PatchTST}} & \multicolumn{2}{c}{\textbf{PhaseFormer}} & \multicolumn{2}{c}{\textbf{TQNet}} & \multicolumn{2}{c}{\textbf{TimeAlign}} & \multicolumn{2}{c}{\textbf{iTransformer}} \\
    \cmidrule(lr){3-4}\cmidrule(lr){5-6}\cmidrule(lr){7-8}\cmidrule(lr){9-10}\cmidrule(lr){11-12}\cmidrule(lr){13-14}\cmidrule(lr){15-16}\cmidrule(lr){17-18}\cmidrule(lr){19-20}\cmidrule(lr){21-22}
    & & MSE & MAE & MSE & MAE & MSE & MAE & MSE & MAE & MSE & MAE & MSE & MAE & MSE & MAE & MSE & MAE & MSE & MAE & MSE & MAE \\
    \midrule
    \multirow{5}{*}{\rotatebox{90}{\textbf{ECL}}} & \textbf{48} & 0.112 & 0.207 & 0.125 & 0.223 & 0.113 & 0.207 & 0.135 & 0.231 & 0.114 & 0.211 & 0.122 & 0.228 & 0.146 & 0.241 & \best{0.110} & \second{0.206} & 0.115 & 0.211 & \second{0.110} & \best{0.205} \\
    & \textbf{96} & \best{0.129} & \best{0.222} & 0.141 & 0.237 & 0.135 & 0.227 & 0.147 & 0.242 & 0.132 & 0.227 & 0.139 & 0.242 & 0.155 & 0.248 & 0.131 & 0.226 & \second{0.130} & \second{0.224} & 0.133 & 0.227 \\
    & \textbf{192} & \best{0.146} & \best{0.239} & 0.155 & 0.249 & 0.151 & 0.244 & 0.174 & 0.270 & \second{0.150} & \second{0.244} & 0.156 & 0.257 & 0.167 & 0.258 & 0.154 & 0.248 & 0.152 & 0.245 & 0.153 & 0.250 \\
    & \textbf{336} & \best{0.164} & \best{0.257} & 0.172 & 0.265 & 0.172 & 0.262 & 0.178 & 0.271 & \second{0.167} & \second{0.261} & 0.173 & 0.274 & 0.183 & 0.273 & 0.171 & 0.265 & 0.169 & 0.263 & 0.174 & 0.267 \\
    \cmidrule(lr){2-22}
    & \textbf{AVG} & \best{0.138} & \best{0.231} & 0.148 & 0.243 & 0.143 & \second{0.235} & 0.159 & 0.254 & \second{0.141} & 0.235 & 0.147 & 0.250 & 0.163 & 0.255 & 0.141 & 0.236 & 0.141 & 0.236 & 0.143 & 0.237 \\
    \midrule
    \multirow{5}{*}{\rotatebox{90}{\textbf{ETTh1}}} & \textbf{48} & \second{0.339} & \best{0.370} & 0.343 & 0.376 & 0.356 & 0.387 & 0.343 & \second{0.374} & 0.382 & 0.412 & 0.346 & 0.386 & 0.363 & 0.387 & \best{0.337} & 0.377 & 0.343 & 0.378 & 0.378 & 0.408 \\
    & \textbf{96} & \second{0.370} & \second{0.391} & 0.376 & 0.397 & 0.395 & 0.411 & \best{0.366} & \best{0.388} & 0.434 & 0.437 & 0.379 & 0.405 & 0.383 & 0.395 & 0.387 & 0.407 & 0.384 & 0.405 & 0.438 & 0.438 \\
    & \textbf{192} & 0.432 & 0.435 & \best{0.405} & \best{0.412} & 0.409 & 0.419 & \second{0.407} & \second{0.413} & 0.444 & 0.446 & 0.432 & 0.433 & 0.419 & 0.419 & 0.410 & 0.418 & 0.429 & 0.433 & 0.509 & 0.475 \\
    & \textbf{336} & 0.473 & 0.461 & \second{0.429} & \second{0.428} & 0.438 & 0.442 & \best{0.418} & \best{0.420} & 0.471 & 0.464 & 0.445 & 0.443 & 0.435 & 0.428 & 0.432 & 0.433 & 0.450 & 0.446 & 0.509 & 0.486 \\
    \cmidrule(lr){2-22}
    & \textbf{AVG} & 0.404 & 0.414 & \second{0.388} & \second{0.403} & 0.400 & 0.415 & \best{0.383} & \best{0.399} & 0.433 & 0.440 & 0.401 & 0.417 & 0.400 & 0.407 & 0.391 & 0.409 & 0.402 & 0.415 & 0.459 & 0.452 \\
    \midrule
    \multirow{5}{*}{\rotatebox{90}{\textbf{ETTh2}}} & \textbf{48} & \second{0.224} & \second{0.300} & \best{0.217} & \best{0.298} & 0.241 & 0.323 & 0.242 & 0.311 & 0.254 & 0.325 & 0.242 & 0.316 & 0.252 & 0.321 & 0.233 & 0.313 & 0.230 & 0.305 & 0.262 & 0.333 \\
    & \textbf{96} & \second{0.283} & \second{0.343} & \best{0.278} & \best{0.339} & 0.311 & 0.366 & 0.293 & 0.344 & 0.324 & 0.372 & 0.310 & 0.364 & 0.301 & 0.355 & 0.296 & 0.352 & 0.293 & 0.349 & 0.369 & 0.393 \\
    & \textbf{192} & \best{0.333} & \best{0.379} & \second{0.337} & 0.384 & 0.375 & 0.402 & 0.347 & \second{0.381} & 0.408 & 0.432 & 0.365 & 0.410 & 0.351 & 0.388 & 0.357 & 0.394 & 0.369 & 0.394 & 0.455 & 0.435 \\
    & \textbf{336} & \best{0.359} & \second{0.403} & \second{0.370} & 0.411 & 0.387 & 0.424 & 0.374 & \best{0.403} & 0.402 & 0.438 & 0.398 & 0.428 & 0.398 & 0.423 & 0.375 & 0.413 & 0.388 & 0.418 & 0.442 & 0.448 \\
    \cmidrule(lr){2-22}
    & \textbf{AVG} & \best{0.300} & \best{0.356} & \second{0.300} & \second{0.358} & 0.328 & 0.379 & 0.314 & 0.360 & 0.347 & 0.392 & 0.329 & 0.379 & 0.326 & 0.372 & 0.315 & 0.368 & 0.320 & 0.367 & 0.382 & 0.402 \\
    \midrule
    \multirow{5}{*}{\rotatebox{90}{\textbf{ETTm1}}} & \textbf{48} & \best{0.251} & \best{0.309} & 0.275 & 0.328 & 0.277 & 0.335 & 0.285 & 0.337 & 0.285 & 0.340 & 0.265 & 0.325 & 0.579 & 0.492 & 0.254 & \second{0.312} & \second{0.252} & 0.313 & 0.309 & 0.346 \\
    & \textbf{96} & \best{0.284} & \best{0.330} & 0.302 & 0.344 & 0.306 & 0.354 & 0.324 & 0.365 & 0.310 & 0.355 & 0.304 & 0.350 & 0.499 & 0.463 & 0.293 & 0.344 & \second{0.291} & \second{0.340} & 0.340 & 0.376 \\
    & \textbf{192} & \second{0.327} & \best{0.359} & 0.337 & 0.365 & 0.344 & 0.379 & 0.350 & 0.377 & 0.348 & 0.381 & 0.333 & 0.376 & 0.542 & 0.490 & 0.335 & 0.372 & \best{0.324} & \second{0.363} & 0.369 & 0.396 \\
    & \textbf{336} & \second{0.360} & \best{0.380} & 0.374 & 0.386 & 0.372 & 0.397 & 0.391 & 0.404 & 0.377 & 0.399 & 0.367 & 0.396 & 0.541 & 0.492 & 0.371 & 0.391 & \best{0.359} & \second{0.385} & 0.410 & 0.423 \\
    \cmidrule(lr){2-22}
    & \textbf{AVG} & \best{0.305} & \best{0.344} & 0.322 & 0.356 & 0.325 & 0.366 & 0.338 & 0.371 & 0.330 & 0.369 & 0.317 & 0.362 & 0.540 & 0.484 & 0.313 & 0.355 & \second{0.307} & \second{0.350} & 0.357 & 0.385 \\
    \midrule
    \multirow{5}{*}{\rotatebox{90}{\textbf{ETTm2}}} & \textbf{48} & \best{0.124} & \best{0.221} & 0.128 & 0.226 & 0.136 & 0.234 & 0.142 & 0.243 & 0.134 & 0.230 & 0.130 & 0.229 & 0.181 & 0.289 & \second{0.127} & \second{0.223} & 0.130 & 0.224 & 0.144 & 0.247 \\
    & \textbf{96} & \best{0.162} & \best{0.251} & \second{0.164} & \second{0.252} & 0.179 & 0.261 & 0.178 & 0.268 & 0.180 & 0.266 & 0.178 & 0.267 & 0.209 & 0.299 & 0.166 & 0.254 & 0.168 & 0.253 & 0.212 & 0.300 \\
    & \textbf{192} & \best{0.221} & \second{0.292} & \second{0.221} & \best{0.291} & 0.237 & 0.306 & 0.232 & 0.304 & 0.243 & 0.301 & 0.233 & 0.304 & 0.266 & 0.331 & 0.226 & 0.296 & 0.237 & 0.298 & 0.278 & 0.343 \\
    & \textbf{336} & \second{0.276} & \second{0.329} & \best{0.275} & \best{0.325} & 0.296 & 0.351 & 0.286 & 0.338 & 0.281 & 0.330 & 0.292 & 0.343 & 0.308 & 0.361 & 0.285 & 0.332 & 0.285 & 0.333 & 0.326 & 0.373 \\
    \cmidrule(lr){2-22}
    & \textbf{AVG} & \best{0.196} & \best{0.273} & \second{0.197} & \second{0.274} & 0.212 & 0.288 & 0.209 & 0.288 & 0.210 & 0.281 & 0.208 & 0.286 & 0.241 & 0.320 & 0.201 & 0.276 & 0.205 & 0.277 & 0.240 & 0.316 \\
    \midrule
    \multirow{5}{*}{\rotatebox{90}{\textbf{Exchange}}} & \textbf{48} & 0.048 & 0.151 & \best{0.044} & \best{0.143} & 0.047 & 0.153 & 0.058 & 0.173 & 0.052 & 0.159 & 0.047 & 0.154 & 0.058 & 0.172 & 0.053 & 0.161 & \second{0.046} & \second{0.150} & 0.053 & 0.166 \\
    & \textbf{96} & \best{0.084} & \second{0.205} & \second{0.088} & \best{0.205} & 0.097 & 0.223 & 0.102 & 0.227 & 0.099 & 0.221 & 0.095 & 0.219 & 0.103 & 0.232 & 0.100 & 0.223 & 0.103 & 0.228 & 0.110 & 0.239 \\
    & \textbf{192} & 0.200 & 0.328 & 0.192 & \second{0.307} & 0.197 & 0.317 & \second{0.187} & 0.311 & 0.238 & 0.346 & 0.188 & 0.315 & 0.214 & 0.335 & 0.196 & 0.317 & \best{0.180} & \best{0.302} & 0.236 & 0.354 \\
    & \textbf{336} & \best{0.313} & \second{0.424} & 0.356 & 0.428 & 0.395 & 0.456 & \second{0.325} & \best{0.415} & 0.393 & 0.451 & 0.356 & 0.435 & 0.378 & 0.452 & 0.425 & 0.477 & 0.355 & 0.432 & 0.444 & 0.488 \\
    \cmidrule(lr){2-22}
    & \textbf{AVG} & \best{0.161} & \second{0.277} & 0.170 & \best{0.271} & 0.184 & 0.287 & \second{0.168} & 0.281 & 0.196 & 0.294 & 0.172 & 0.281 & 0.188 & 0.298 & 0.193 & 0.294 & 0.171 & 0.278 & 0.211 & 0.312 \\
    \midrule
    \multirow{5}{*}{\rotatebox{90}{\textbf{Traffic}}} & \textbf{48} & 0.383 & 0.260 & 0.394 & 0.275 & 0.347 & 0.252 & 0.402 & 0.276 & 0.363 & 0.263 & 0.352 & 0.256 & 0.414 & 0.264 & \second{0.342} & \best{0.237} & 0.367 & 0.262 & \best{0.329} & \second{0.243} \\
    & \textbf{96} & 0.397 & 0.266 & 0.411 & 0.280 & 0.371 & \second{0.258} & 0.416 & 0.279 & 0.388 & 0.276 & 0.376 & 0.268 & 0.425 & 0.266 & \second{0.364} & \best{0.252} & 0.376 & 0.261 & \best{0.363} & 0.260 \\
    & \textbf{192} & 0.410 & 0.275 & 0.424 & 0.285 & \second{0.393} & \best{0.264} & 0.430 & 0.287 & 0.407 & 0.284 & 0.395 & 0.276 & 0.432 & \second{0.268} & 0.399 & 0.278 & 0.413 & 0.288 & \best{0.385} & 0.270 \\
    & \textbf{336} & 0.428 & 0.285 & 0.436 & 0.291 & 0.422 & 0.293 & 0.443 & 0.295 & 0.421 & 0.291 & \second{0.409} & 0.282 & 0.442 & \best{0.272} & 0.420 & 0.297 & 0.428 & 0.295 & \best{0.392} & \second{0.274} \\
    \cmidrule(lr){2-22}
    & \textbf{AVG} & 0.405 & 0.272 & 0.416 & 0.283 & 0.383 & 0.267 & 0.423 & 0.284 & 0.395 & 0.278 & 0.383 & 0.270 & 0.428 & 0.267 & \second{0.381} & \second{0.266} & 0.396 & 0.276 & \best{0.367} & \best{0.262} \\
    \midrule
    \multirow{5}{*}{\rotatebox{90}{\textbf{Weather}}} & \textbf{48} & 0.116 & 0.160 & 0.136 & 0.184 & 0.120 & 0.169 & 0.146 & 0.200 & 0.118 & 0.163 & \best{0.114} & \best{0.155} & 0.153 & 0.207 & 0.120 & 0.163 & \second{0.116} & \second{0.157} & 0.131 & 0.176 \\
    & \textbf{96} & \best{0.145} & \best{0.196} & 0.174 & 0.224 & 0.155 & 0.209 & 0.183 & 0.236 & 0.151 & 0.204 & 0.150 & \second{0.198} & 0.184 & 0.244 & 0.156 & 0.209 & \second{0.149} & 0.199 & 0.167 & 0.221 \\
    & \textbf{192} & \best{0.188} & \best{0.237} & 0.217 & 0.259 & 0.194 & 0.242 & 0.225 & 0.269 & 0.197 & 0.250 & 0.197 & 0.244 & 0.225 & 0.273 & 0.202 & 0.250 & \second{0.189} & \second{0.237} & 0.207 & 0.254 \\
    & \textbf{336} & \best{0.242} & \best{0.278} & 0.264 & 0.293 & 0.242 & 0.279 & 0.273 & 0.304 & 0.249 & 0.288 & 0.254 & 0.287 & 0.272 & 0.307 & 0.249 & 0.284 & \second{0.242} & \second{0.279} & 0.257 & 0.291 \\
    \cmidrule(lr){2-22}
    & \textbf{AVG} & \best{0.173} & \best{0.218} & 0.198 & 0.240 & 0.178 & 0.225 & 0.207 & 0.252 & 0.179 & 0.226 & 0.179 & 0.221 & 0.208 & 0.257 & 0.182 & 0.226 & \second{0.174} & \second{0.218} & 0.191 & 0.235 \\
    \bottomrule
  \end{tabular}}
\end{table}

\subsection{Forecasting Visualizations}
\label{app:forecast-examples}

\begin{figure}[p]
  \centering
  \includegraphics[width=\linewidth]{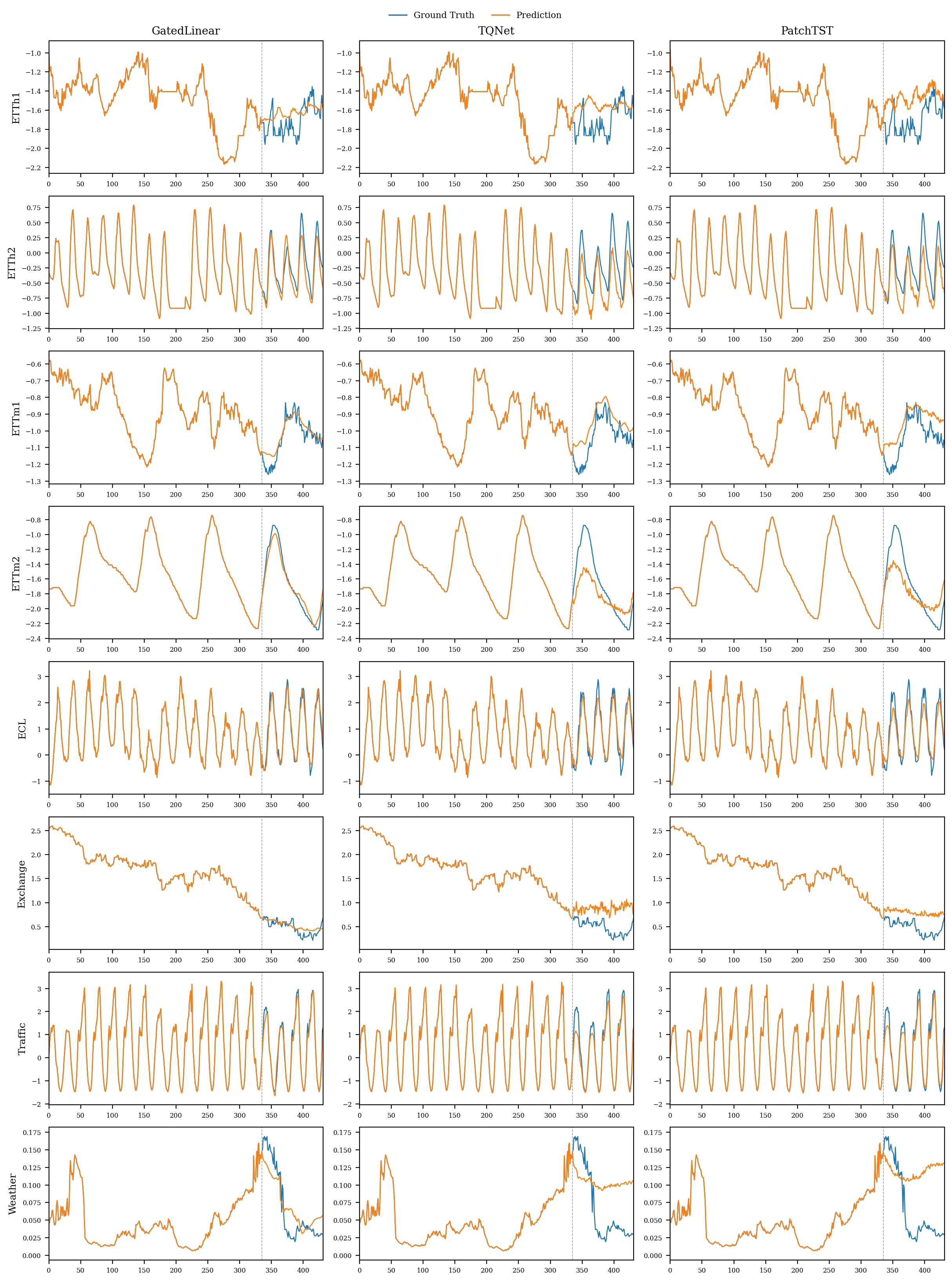}
  \caption{Forecasting visualization examples on eight benchmark datasets with input length $L=336$ and prediction length $H=96$. Each panel compares the predicted trajectory of our model with representative baselines against the ground truth, providing a qualitative view of how GatedLinear tracks diverse temporal patterns across datasets.}
  \label{fig:appendix-forecast-examples}
\end{figure}

\begin{figure}[p]
  \centering
  \includegraphics[width=\linewidth]{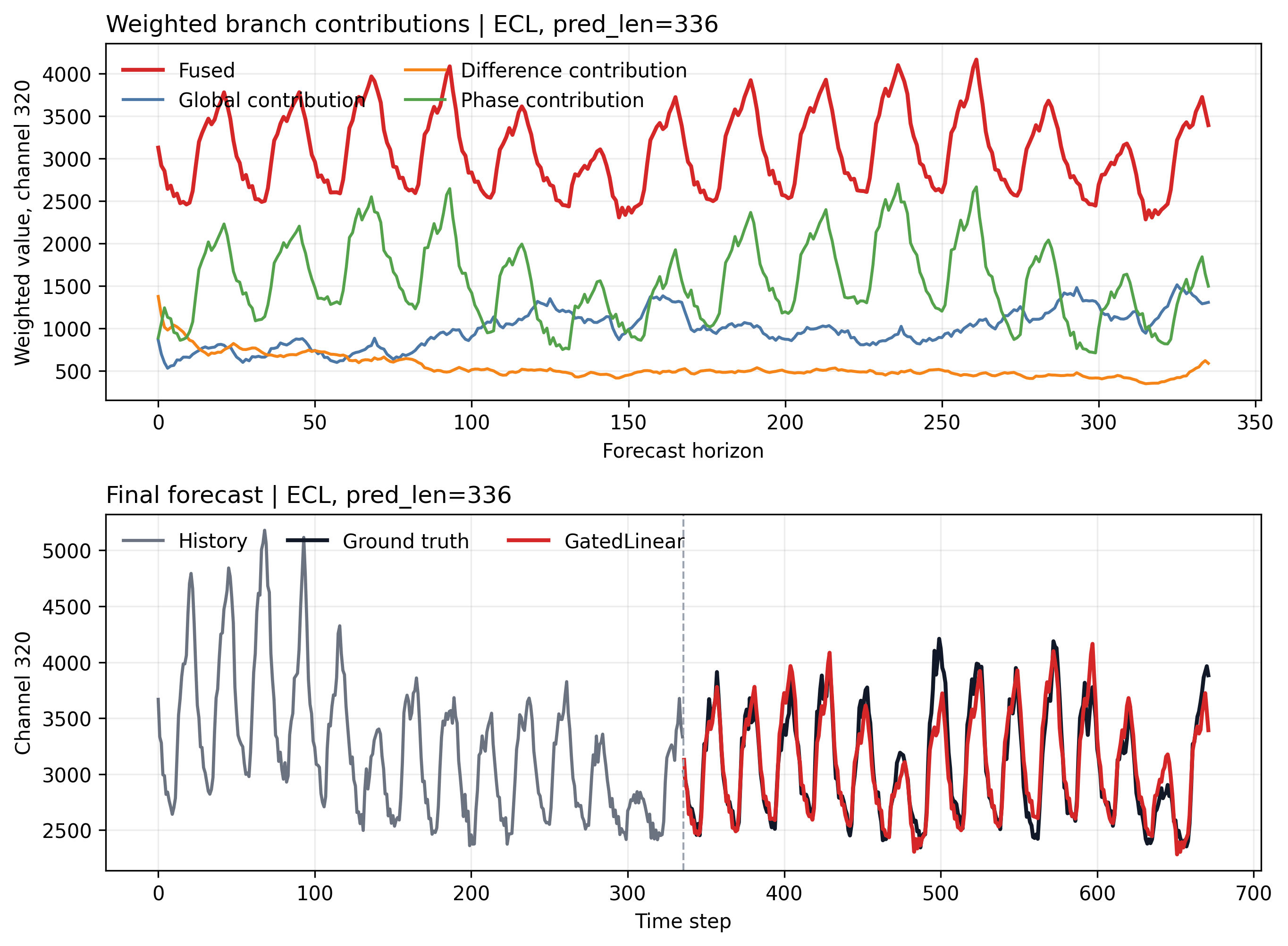}
  \vspace{0.75em}
  \includegraphics[width=\linewidth]{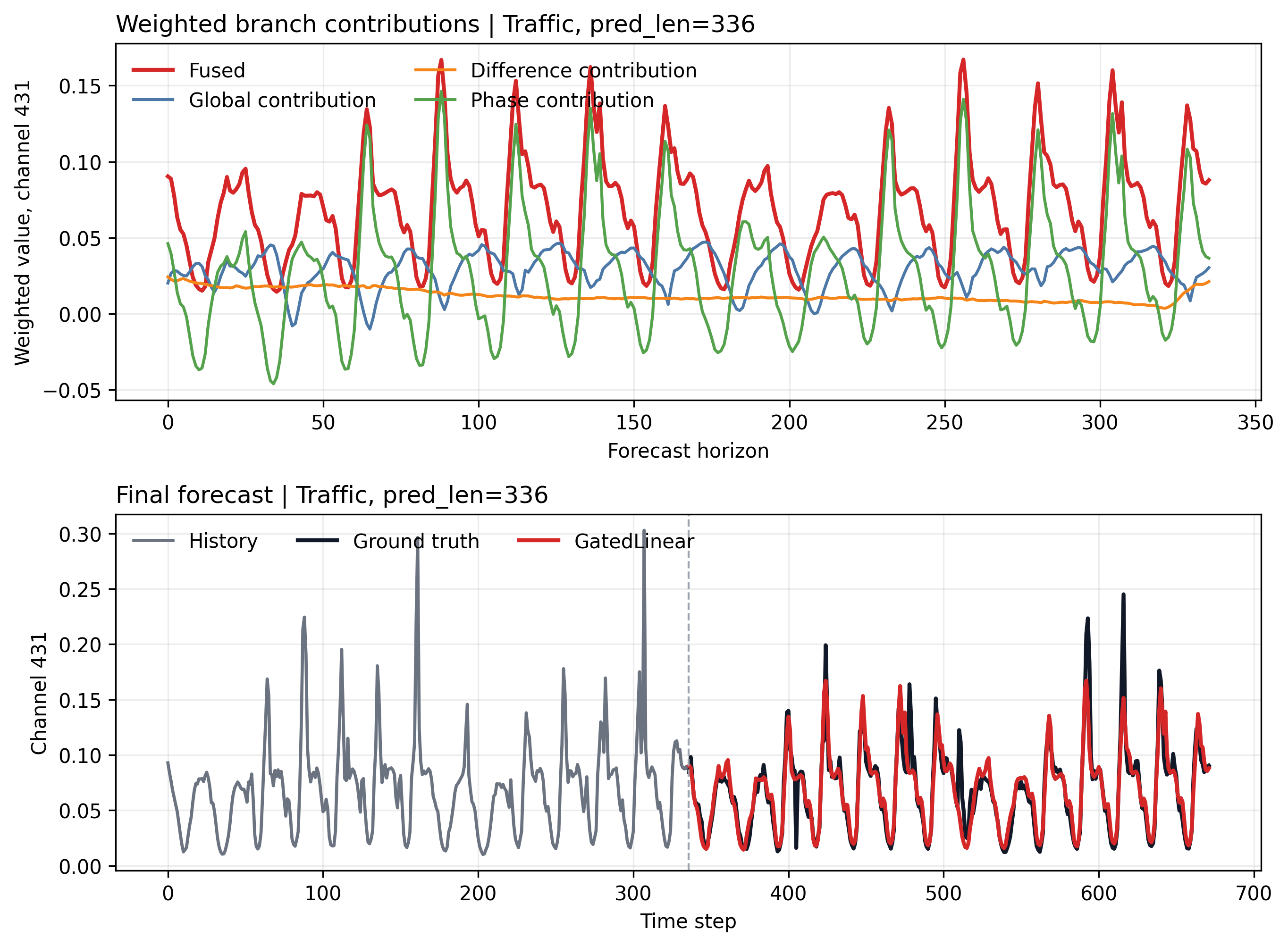}
  \caption{Forecasting visualizations with weighted tri-branch contributions on ECL and Traffic, using input length $L=336$ and prediction length $H=336$. For each dataset, the top panel shows the fused prediction together with the weighted contributions from the global, difference-based, and phase-aligned branches, and the bottom panel compares the final forecast with the ground truth.}
  \label{fig:appendix-branch-contribution-forecasts}
\end{figure}

Figure~\ref{fig:appendix-forecast-examples} provides qualitative examples for the same forecasting setting. The vertical dashed line separates the observed input window from the prediction region. Across strongly periodic examples such as ECL and Traffic, GatedLinear preserves the phase and amplitude of repeated peaks more closely than the shown baselines, which is consistent with the role of the phase-aligned branch. On ETTm2 and ETTh2, the predictions also follow the dominant future shape while avoiding large phase shifts. For more irregular or nonstationary series such as ETTh1, Exchange, and Weather, all methods have difficulty matching abrupt local changes exactly; GatedLinear tends to produce smoother and more conservative trajectories, tracking the main level and trend while not overreacting to short-lived fluctuations. These visualizations complement the quantitative results by showing that the proposed gate can select useful forecasting mechanisms under diverse temporal patterns, while the remaining errors often occur around sudden regime changes or high-frequency variations.

Figure~\ref{fig:appendix-branch-contribution-forecasts} further shows forecasting results with the three branch contributions on ECL and Traffic. The visualizations make the fused forecast and the global, difference-based, and phase-aligned components explicit, illustrating how the final prediction is formed by combining complementary branches over the forecast horizon.

\subsection{Effect of Different Input Windows}
\label{app:input-window-comparison}

\begin{figure}[p]
  \centering
  \includegraphics[width=\linewidth]{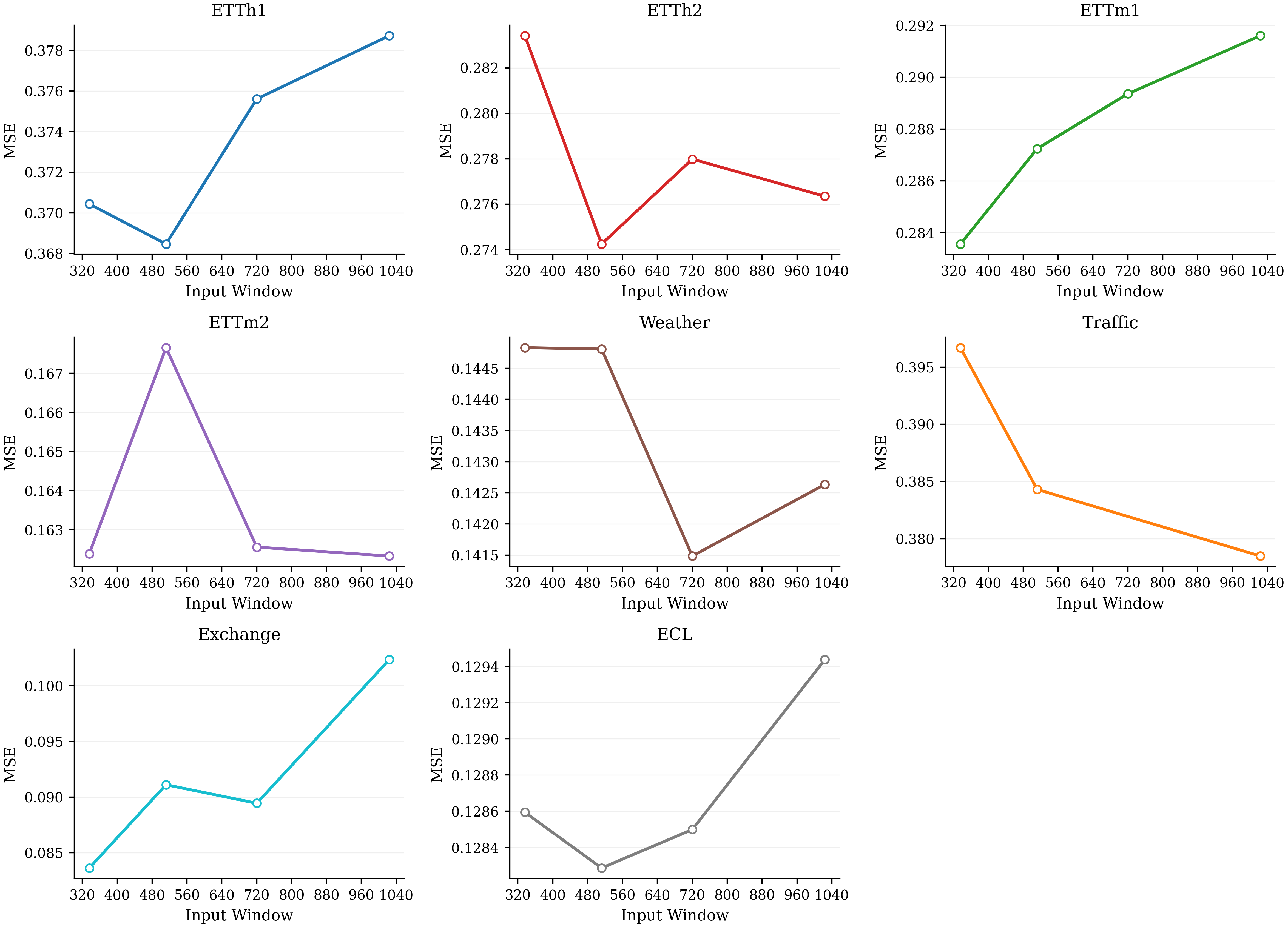}
  \caption{Performance comparison of GatedLinear under different input lengths across benchmark datasets, with prediction length fixed to $H=96$. The results show the effect of using different historical context windows for a fixed forecasting horizon.}
  \label{fig:appendix-input-window-comparison}
\end{figure}

Figure~\ref{fig:appendix-input-window-comparison} compares GatedLinear under different input-window lengths and examines whether longer input histories always improve forecasting accuracy. The effect is dataset-dependent. Traffic benefits from a longer window, and Weather also improves when more historical context is provided, which is plausible because both datasets contain stable daily or weekly patterns that can be reinforced by additional cycles. In contrast, ETTm1 and Exchange become worse when the window is excessively long, suggesting that older observations may introduce stale or nonstationary information. ETTh1, ETTh2, ETTm2, and ECL show their best or near-best results at moderate window lengths, rather than at the maximum window. Overall, the comparison indicates that the useful context size depends on the balance between informative periodic evidence and outdated historical information.

\subsection{Parameter Sensitivity}
\label{app:parameter-sensitivity}

\begin{figure}[p]
  \centering
  \includegraphics[width=\linewidth]{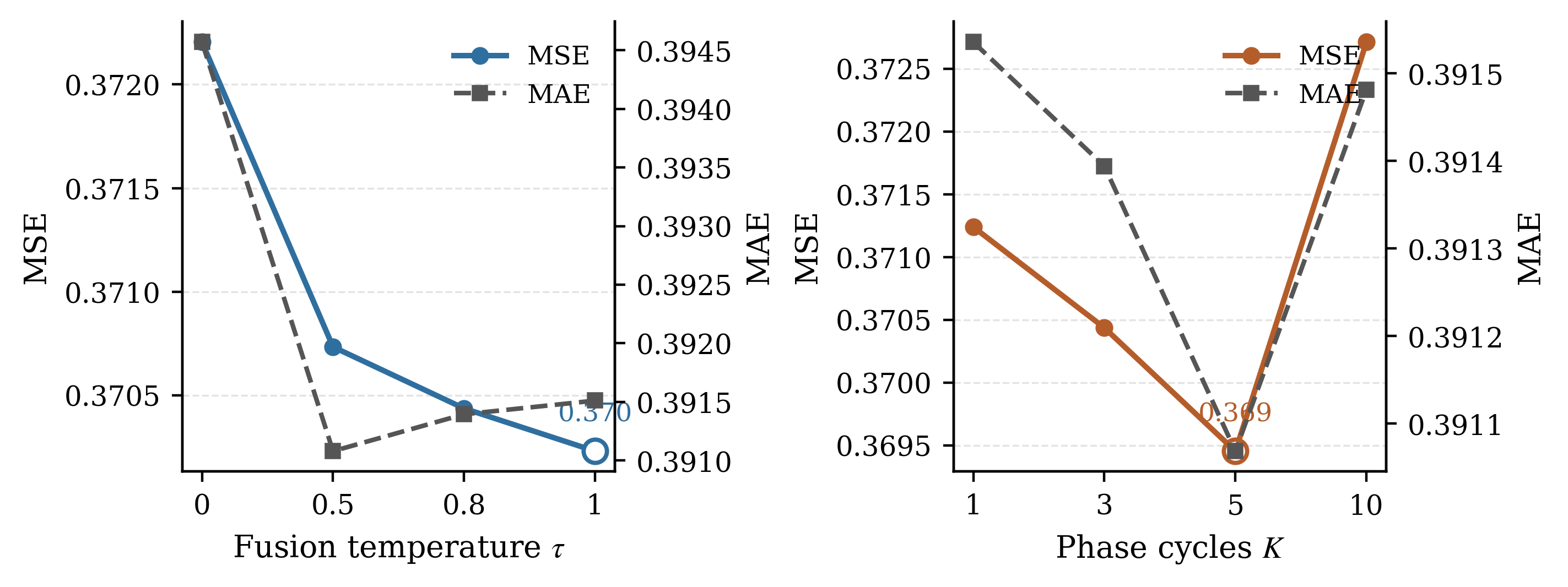}
  \caption{Parameter sensitivity study on the ETTh1 dataset. The left panel varies the fusion temperature $\tau$ in the fusion gate, and the right panel varies the number of phase cycles $K$. MSE and MAE are reported to show how these two hyperparameters affect forecasting performance.}
  \label{fig:appendix-parameter-sensitivity}
\end{figure}

Figure~\ref{fig:appendix-parameter-sensitivity} analyzes two hyperparameters of the proposed gate and phase branch. For the fusion temperature $\tau$, very sharp routing gives higher error, while moderate temperatures lead to lower MSE and MAE. The performance remains close across $\tau\in[0.5,1.0]$, and the default value $\tau=0.8$ lies in this stable low-error region, indicating that the model does not rely on a finely tuned temperature. For the number of phase cycles $K$, using a small number of recent cycles is beneficial, with the best region around $K=3$ to $K=5$ in this study. Increasing $K$ to 10 degrades both metrics, likely because older cycles dilute the current phase template when the series undergoes level or amplitude changes. These results support the design choice of using compact phase evidence and a softly adaptive fusion gate.

\clearpage

\newpage
\section*{NeurIPS Paper Checklist}

\begin{enumerate}

\item {\bf Claims}
    \item[] Question: Do the main claims made in the abstract and introduction accurately reflect the paper's contributions and scope?
    \item[] Answer: \answerYes{}
    \item[] Justification: The abstract and introduction state the proposed tri-basis GatedLinear framework, its Tri-Factorized Fusion Gate, interpretability, and efficiency claims. These claims are evaluated through the methodology, main results, ablations, interpretability analysis, and efficiency analysis.

\item {\bf Limitations}
    \item[] Question: Does the paper discuss the limitations of the work performed by the authors?
    \item[] Answer: \answerYes{}
    \item[] Justification: The appendix includes a dedicated limitations section discussing the channel-independent design, dependence on meaningful periods and future phase indices, and the limited expressiveness of the three linear forecasting bases.

\item {\bf Theory assumptions and proofs}
    \item[] Question: For each theoretical result, does the paper provide the full set of assumptions and a complete (and correct) proof?
    \item[] Answer: \answerYes{}
    \item[] Justification: Appendix~\ref{app:theoretical-analysis} states the assumptions for a non-stationary integrated random walk with drift, derives the corresponding MMSE predictor, and analyzes why a global basis is not shift-invariant under this setting. The same appendix shows that the difference-based basis recovers the MMSE predictor structure and remains invariant to arbitrary level shifts under the stated assumptions.

    \item {\bf Experimental result reproducibility}
    \item[] Question: Does the paper fully disclose all the information needed to reproduce the main experimental results of the paper to the extent that it affects the main claims and/or conclusions of the paper (regardless of whether the code and data are provided or not)?
    \item[] Answer: \answerYes{}
    \item[] Justification: The paper and supplemental material describe the architecture, datasets, splits, metrics, optimizer, evaluated horizons, random-seed protocol, and hyperparameter settings used to obtain the main results. The supplemental code package provides the implementation and reproduction instructions for the reported experiments.

\item {\bf Open access to data and code}
    \item[] Question: Does the paper provide open access to the data and code, with sufficient instructions to faithfully reproduce the main experimental results, as described in supplemental material?
    \item[] Answer: \answerYes{}
    \item[] Justification: The experiments use standard public forecasting benchmarks, and the supplemental material includes an anonymized code package with instructions for reproducing the main results. The authors plan to release a public repository after the review process.

\item {\bf Experimental setting/details}
    \item[] Question: Does the paper specify all the training and test details (e.g., data splits, hyperparameters, how they were chosen, type of optimizer) necessary to understand the results?
    \item[] Answer: \answerYes{}
    \item[] Justification: The experimental setup specifies the benchmark datasets, train/validation/test split, input and prediction lengths, metrics, optimizer, validation-based hyperparameter selection, and random-seed averaging. Additional hyperparameter and implementation details are provided in the appendix and supplemental material.

\item {\bf Experiment statistical significance}
    \item[] Question: Does the paper report error bars suitably and correctly defined or other appropriate information about the statistical significance of the experiments?
    \item[] Answer: \answerNo{}
    \item[] Justification: Reported numbers are averaged over three random seeds, and the main conclusions are supported by mean performance across multiple datasets, horizons, and ablations. Explicit standard deviations, confidence intervals, or significance tests are not included due to space constraints.

\item {\bf Experiments compute resources}
    \item[] Question: For each experiment, does the paper provide sufficient information on the computer resources (type of compute workers, memory, time of execution) needed to reproduce the experiments?
    \item[] Answer: \answerYes{}
    \item[] Justification: The paper reports the GPU information used for the experiments and includes an efficiency analysis comparing training time, forecasting accuracy, and peak GPU memory. These details provide the compute context needed to reproduce and interpret the reported experiments.
    
\item {\bf Code of ethics}
    \item[] Question: Does the research conducted in the paper conform, in every respect, with the NeurIPS Code of Ethics \url{https://neurips.cc/public/EthicsGuidelines}?
    \item[] Answer: \answerYes{}
    \item[] Justification: The work is a forecasting-method study using standard benchmark time-series datasets and does not involve human-subject experiments, private data collection, or high-risk model release. The research conforms to the NeurIPS Code of Ethics.

\item {\bf Broader impacts}
    \item[] Question: Does the paper discuss both potential positive societal impacts and negative societal impacts of the work performed?
    \item[] Answer: \answerNo{}
    \item[] Justification: The paper does not include a separate broader-impact discussion because it presents a general methodological contribution for time series forecasting rather than an application-specific deployed system. The work is evaluated on standard public benchmarks and aims to improve the accuracy, efficiency, and interpretability of forecasting models; any societal benefits or risks would primarily depend on the downstream domain in which such a forecaster is deployed.
    
\item {\bf Safeguards}
    \item[] Question: Does the paper describe safeguards that have been put in place for responsible release of data or models that have a high risk for misuse (e.g., pre-trained language models, image generators, or scraped datasets)?
    \item[] Answer: \answerNA{}
    \item[] Justification: The paper does not release high-risk models or scraped datasets, and the proposed forecasting architecture is evaluated on standard benchmark time-series data. No special controlled-release safeguards are therefore applicable.

\item {\bf Licenses for existing assets}
    \item[] Question: Are the creators or original owners of assets (e.g., code, data, models), used in the paper, properly credited and are the license and terms of use explicitly mentioned and properly respected?
    \item[] Answer: \answerYes{}
    \item[] Justification: The paper properly credits the benchmark datasets and baseline methods through citations, and the appendix provides the official repositories for the compared baselines. The experiments use public benchmark assets and baseline implementations according to their original licenses and terms of use, with additional asset and license information documented in the supplemental material where applicable.

\item {\bf New assets}
    \item[] Question: Are new assets introduced in the paper well documented and is the documentation provided alongside the assets?
    \item[] Answer: \answerYes{}
    \item[] Justification: The paper introduces GatedLinear and provides an anonymized supplemental code package for the proposed model. The released code is documented with instructions for environment setup, data preparation, training, evaluation, hyperparameter settings, and reproduction of the main experiments; a public repository will be released after the review process.

\item {\bf Crowdsourcing and research with human subjects}
    \item[] Question: For crowdsourcing experiments and research with human subjects, does the paper include the full text of instructions given to participants and screenshots, if applicable, as well as details about compensation (if any)? 
    \item[] Answer: \answerNA{}
    \item[] Justification: The paper does not involve crowdsourcing, participant studies, or other research with human subjects. All experiments are conducted on time-series forecasting benchmarks.

\item {\bf Institutional review board (IRB) approvals or equivalent for research with human subjects}
    \item[] Question: Does the paper describe potential risks incurred by study participants, whether such risks were disclosed to the subjects, and whether Institutional Review Board (IRB) approvals (or an equivalent approval/review based on the requirements of your country or institution) were obtained?
    \item[] Answer: \answerNA{}
    \item[] Justification: The work does not include human-subject research or crowdsourcing experiments. IRB or equivalent approval is therefore not applicable to the described study.

\item {\bf Declaration of LLM usage}
    \item[] Question: Does the paper describe the usage of LLMs if it is an important, original, or non-standard component of the core methods in this research? Note that if the LLM is used only for writing, editing, or formatting purposes and does \emph{not} impact the core methodology, scientific rigor, or originality of the research, declaration is not required.
    \item[] Answer: \answerNA{}
    \item[] Justification: The core method is a linear time-series forecasting architecture with a structured fusion gate and does not use LLMs as an original or non-standard methodological component.

\end{enumerate}

\end{document}